%% file: main.tex
\documentclass[acmtog,nonacm]{acmart}

\usepackage{fp}
\usepackage{expl3}[2012-07-08]
\ExplSyntaxOn
\ExplSyntaxOff

\usepackage{amsmath,amsfonts,bm}
\usepackage{bm}

\def\x{{x}}

\def\xi{{\x_i}}

\newcommand{\ignorethis}[1]{}

\def\eqref#1{equation~\ref{#1}}

\def\1{\bm{1}}

\DeclareMathAlphabet{\mathsfit}{\encodingdefault}{\sfdefault}{m}{sl}
\SetMathAlphabet{\mathsfit}{bold}{\encodingdefault}{\sfdefault}{bx}{n}

\newcommand{\ignore}[1]{}

\makeatletter
\DeclareRobustCommand\onedot{\futurelet\@let@token\@onedot}
\def\@onedot{\ifx\@let@token.\else.\null\fi\xspace}

\makeatother
\newcommand{\ie}{\textit{i}.\textit{e}.}
\newcommand{\eg}{\textit{e}.\textit{g}.}

\usepackage{multirow}
\usepackage{booktabs} 
\usepackage{url}
\usepackage{stfloats}

\usepackage{enumitem}
\setlist[itemize]{align=parleft,left=0pt..1em}
\usepackage{listings}
\usepackage{amsmath,bm}
\usepackage[ruled]{algorithm2e} 

\SetAlFnt{\small}
\SetAlCapFnt{\small}
\SetAlCapNameFnt{\small}
\SetAlCapHSkip{0pt}
\setcopyright{none}
\usepackage{overpic} 
\usepackage{subcaption}

\DeclareMathAlphabet{\mathcal}{OMS}{cmsy}{m}{n}

\usepackage{array}
\newcolumntype{L}[1]{>{\raggedright\let\newline\\\arraybackslash\hspace{0pt}}m{#1}}
\newcolumntype{C}[1]{>{\centering\let\newline\\\arraybackslash\hspace{0pt}}m{#1}}
\newcolumntype{R}[1]{>{\raggedleft\let\newline\\\arraybackslash\hspace{0pt}}m{#1}}

\newcommand{\sysName}{LayerPeeler}

\begin{document}
\begin{sloppypar}
\title{LayerPeeler: Autoregressive Peeling for Layer-wise Image Vectorization}

\author{Ronghuan Wu}
\affiliation{
  \institution{City University of Hong Kong}
  \country{China}
}
\email{rh.wu@my.cityu.edu.hk}
\author{Wanchao Su}
\affiliation{
  \institution{Monash University}
  \country{Australia}
}
\email{wanchao.su@monash.edu}
\author{Jing Liao}
\authornote{Corresponding Author}
\affiliation{
  \institution{City University of Hong Kong}
  \country{China}
}
\email{jingliao@cityu.edu.hk}

\begin{abstract}
Image vectorization is a powerful technique that converts raster images into vector graphics, enabling enhanced flexibility and interactivity.
However, popular image vectorization tools struggle with occluded regions, producing incomplete or fragmented shapes that hinder editability.
While recent advancements have explored optimization-based and learning-based layer-wise image vectorization, these methods face limitations in vectorization quality and flexibility.
In this paper, we introduce \sysName, a novel layer-wise image vectorization approach that addresses these challenges through a progressive simplification paradigm.
The key to \sysName's success lies in its autoregressive peeling strategy: by identifying and removing the topmost non-occluded layers while recovering underlying content, we generate vector graphics with complete paths and coherent layer structures.
Our method leverages vision-language models to construct a layer graph that captures occlusion relationships among elements, enabling precise detection and description for non-occluded layers.
These descriptive captions are used as editing instructions for a finetuned image diffusion model to remove the identified layers.
To ensure accurate removal, we employ localized attention control that precisely guides the model to target regions while faithfully preserving the surrounding content.
To support this, we contribute a large-scale dataset specifically designed for layer peeling tasks.
Extensive quantitative and qualitative experiments demonstrate that \sysName~significantly outperforms existing techniques, producing vectorization results with superior path semantics, geometric regularity, and visual fidelity.
Our code and dataset will be available at {\color{blue}\url{https://layerpeeler.github.io/}}.
\end{abstract}

%
%

\begin{teaserfigure}
  \centering
  \includegraphics[width=\textwidth]{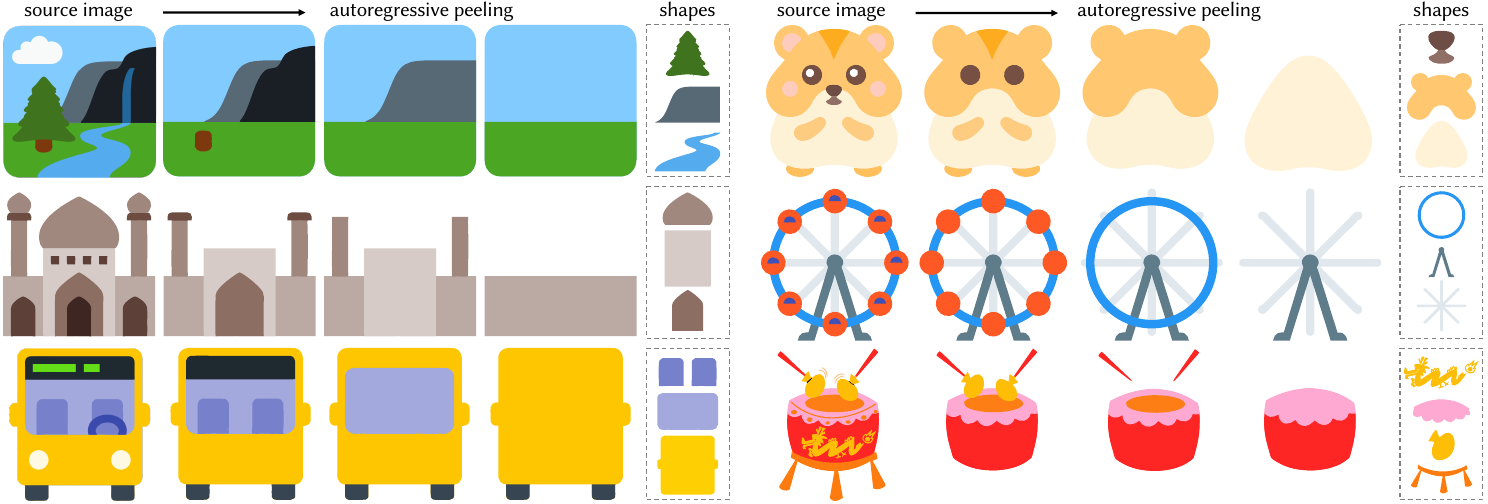}
  \caption{\sysName~enables a layer-wise image vectorization that autoregressively removes topmost layers to recover occluded regions. Each example showcases the key steps in the peeling process, with representative shapes highlighted in the box to demonstrate the structural coherence of the recovered layers. The icon images are from \copyright SVGRepo.
  }
  \label{fig:teaser}
\end{teaserfigure}


\maketitle

\section{Introduction}
Vector graphics encode visual content through mathematically defined geometric primitives, offering resolution-independent rendering and explicit hierarchical representation via layered structures.
Unlike raster images, vector representations enable lossless geometric transformations and precise element-wise manipulation, making them widely used in 3D rendering, digital art, and interface design, where structural fidelity and parametric control are essential.

\begin{figure}[t]
    \centering
    \includegraphics[width=0.9\linewidth]{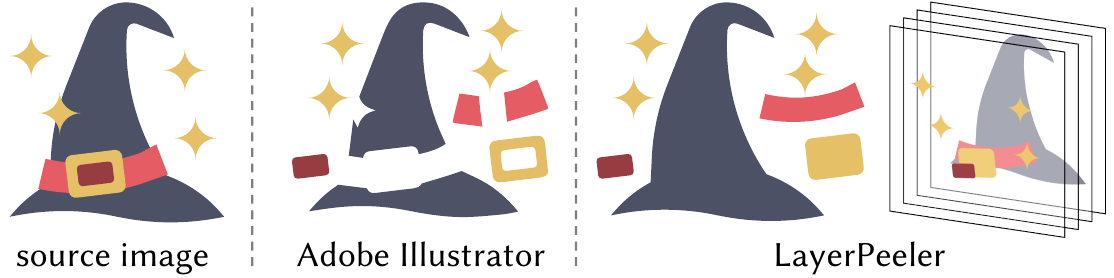}
    \caption{Existing vectorization tools (\eg, Adobe Illustrator) produce vectorizations with incomplete or fragmented shapes. 
    In contrast, our \sysName~produces layer-wise vector elements that preserve layer hierarchies, maintain structural integrity, and exhibit clear semantic correspondence.
    The icon image is from \copyright SVGRepo.
    }
\label{fig:vectorization}
\end{figure}

Manually creating complex vector graphics content is a challenging task that demands a steep learning curve and professional software expertise. 
Image vectorization provides an automated solution to convert raster images into vector formats with minimal manual intervention.
However, existing \textit{rule-based} algorithms~\cite{tian2022survey} and popular vectorization tools (\eg, Potrace and Adobe Illustrator) suffer from a significant limitation: when processing a partially occluded color region, they often produce incomplete or fragmented vector elements (see Fig.~\ref{fig:vectorization}).
These disjointed components impede intuitive shape manipulation and may introduce visual artifacts when transformed, as the intended semantic structures are not preserved.
To address these issues, a layer-aware vectorization method is necessary -- one that can reconstruct occluded regions and generate well-structured, editable vector layers while maintaining topological coherence.

To this end, one line of work pursues \textit{optimization-based} geometric methods~\cite{favreau2017photo2clipart,du2023image,law2025image}. Starting from a segmentation map, they optimize an objective that trades off image fidelity against structural regularization (\eg, layer count and smooth boundaries) and infer layer order from perceptual cues (\eg, T-/X-junctions) or energy-based criteria. However, with shape completion based on geometric features instead of semantics, reconstructions are frequently oversimplified and sometimes fail entirely.
Beyond purely geometric formulations, a second optimization-based line leverages differentiable rendering to optimize vector primitives~\cite{ma2022towards,hirschorn2024optimize,zhou2024segmentation,wang2024layered}. These methods use a differentiable rasterizer~\cite{Li:2020:DVG} to minimize reconstruction or perceptual losses, and synthesize layers jointly or progressively from background (deepest) to foreground, which can yield detailed reconstructions but often over-partition shapes and produce messy layer structures.
In parallel, \textit{learning-based} methods~\cite{reddy2021im2vec,shen2021clipgen,chen2023editable,thamizharasan2024vecfusion,song2025layertracer} aim to predict layered vectors in a single pass, but are bottlenecked by the scarcity of large-scale vector datasets, limiting generalization across styles and complexities.

In this paper, we develop a learning-based image vectorization method that converts raster images into vector graphics with strong structural integrity through an iterative, layer-wise approach.
The key to the success of \sysName~lies in its autoregressive peeling paradigm, which progressively identifies and removes the topmost non-occluded layers while recovering the occluded content beneath -- hence the name \sysName.
Our method leverages the spatial understanding capabilities of Vision-Language Models (VLMs) to infer occlusion relationships and layer ordering, enabling precise detection and description of non-occluded layers.
To mitigate the scarcity of vector graphics data, we harness the strong image priors of image diffusion models. Specifically, we employ a diffusion transformer model fine-tuned with Low-Rank Adaptation (LoRA)~\cite{hu2021lora} to interpret layer descriptions and synthesize clean removals of targeted elements.
To further enhance visual fidelity and editing precision, we introduce a localized attention control mechanism that constrains modifications to relevant regions during denoising, ensuring seamless integration with the surrounding context for fine-grained element removal.

We demonstrate the effectiveness of our method in the context of Scalable Vector Graphics (SVG), a widely used vector image format.
To support the layer peeling task, we construct a new dataset from a large collection of SVG collected from the Internet, enriched with detailed annotations of the topmost layers.
This dataset is used to train our image diffusion model for accurate layer-wise modification.
We conduct a comprehensive evaluation of \sysName~across path semantics, geometric regularity, and visual fidelity.
Our experimental results show that \sysName~consistently outperforms existing optimization-based and learning-based methods both quantitatively and qualitatively. 
Moreover, we demonstrate the method's generalization capability across varying SVG styles and complexity levels.
Our main contributions are summarized as follows:
\begin{itemize}[leftmargin=*]
\item We introduce a novel layer-wise image vectorization framework
that combines VLMs' reasoning strengths with image diffusion models' editing capabilities.
\item We exploit the spatial reasoning power of VLMs to construct a layer graph, enabling accurate detection of non-occluded regions.
\item We develop a precise and fidelity-preserving region removal strategy by controlling the attention mask in image diffusion models.
\item We construct a new large-scale dataset with detailed layer annotations, tailored for fine-grained layer peeling tasks.
\end{itemize}

\section{Related Work}
In this section, we first review prior work in image vectorization (Section~\ref{subsec:image_vectorization}), which forms the foundation of our approach.
We then discuss relevant advances in vision-language models for spatial understanding (Section~\ref{subsec:vlm_reasoning}) and diffusion models for image editing (Section~\ref{subsec:diffusion_editing}), which are key components that enable our layer-wise vectorization framework.

\subsection{Image Vectorization}
\label{subsec:image_vectorization}
Early work on image vectorization predominantly focused on \textit{rule-based} approaches, which laid the groundwork for vector graphics representations and creative tools.
These methods typically represented vector graphics using meshes~\cite{battiato2004svg, demaret2006image, swaminarayan2006rapid, lecot2006ardeco, liao2012subdivision, xia2009patch, yang2015effective, sun2007image} or curves~\cite{orzan2008diffusion, xie2014hierarchical, zhao2017inverse}, offering high-quality vectorization of visual content.
However, a key limitation of these algorithmic techniques is their inability to recover occluded content, often resulting in incomplete or fragmented shapes. We refer readers to~\cite{tian2022survey} for a more comprehensive review of these foundational approaches.

A critical challenge in enabling layer-wise image vectorization is accurately determining the layer order while recovering the occluded content.
To this end, \textit{optimization-based} methods cast vectorization as energy minimization that trades off reconstruction fidelity against structural regularity.
Within this family, geometric optimization approaches follow a common recipe: starting from a segmentation map, they formulate an objective that explicitly balances a reconstruction term against structural regularizers (\eg, penalizing layer count, encouraging semi-transparency and convex/smooth boundaries), infer layer order from perceptual cues (\eg, T-/X-junctions) or energy-based criteria, and complete occlusions via geometric or variational models, yielding a small number of editable layers~\cite{favreau2017photo2clipart,entem2018structuring,du2023image,law2025image}. However, because completion relies on geometric features rather than semantics, these methods can oversimplify occluded regions, struggle with non-convex objects, and are sensitive to segmentation quality, leading to semantically implausible layers or orderings in complex scenes.
In parallel, differentiable rendering optimization methods iteratively refine vector graphics by minimizing reconstruction or perceptual losses through a differentiable rasterizer~\cite{Li:2020:DVG}, achieving high visual fidelity but frequently at the cost of an excessive number of layers and fragmented or misaligned regions~\cite{ma2022towards,hirschorn2024optimize,zhou2024segmentation,wang2024layered}.

More recently, \textit{learning-based} methods leverage deep networks to directly map raster images to vector representations.
By learning depth priors and structural patterns from data, these models enable single-pass vectorization~\cite{reddy2021im2vec, shen2021clipgen, chen2023editable, thamizharasan2024vecfusion, song2025layertracer}.
However, their performance remains constrained by the limited availability of large-scale vector graphics datasets, which hinders generalization across diverse styles and domains.

Our work focuses on vectorizing non-photorealistic images~\cite{dominici2020polyfit, du2023image, favreau2017photo2clipart, hoshyari2018perception, yan2024deep, Zhang2009VectorizingCA, su2021marvel, kopf2011depixelizing}, which play an important role in design applications, through an iterative, layer-wise approach. We overcome previous limitations by leveraging VLMs for sophisticated layer-order analysis and by addressing data scarcity through the rich prior knowledge embedded in image diffusion models.

\subsection{Vision-Language Models for Spatial Understanding}
\label{subsec:vlm_reasoning}
Vision-language models have demonstrated remarkable capabilities in understanding and reasoning about visual content, enabling diverse applications across multiple domains. These include visual layout configuration, such as poster design and 3D scene arrangement~\cite{cheng2024graphic, yang2024posterllava, sun2024layoutvlm}, creative content manipulation, like 3D shape editing and motion graphics generation~\cite{ganeshan2024parsel, huang2024blenderalchemy, liu2024logomotion, ma2025mover}, and high-level design understanding and evaluation~\cite{kulits2024re, wu2024gpt}.
Leveraging VLMs' deep understanding of visual concepts, we employ them to analyze depth and occlusion relationships in input images and construct accurate layer graphs describing the topmost elements. Additionally, through carefully designed visual annotations~\cite{yang2023set, cai2024vip, lei2024scaffolding}, we harness their capabilities to annotate layers in vector graphics.

\subsection{Diffusion Models for Image Editing}
\label{subsec:diffusion_editing}
Image editing with diffusion models can be broadly categorized into learning-based and training-free approaches.
learning-based methods~\cite{brooks2023instructpix2pix, wei2024omniedit, xu2024insightedit, yu2024anyedit, zhang2023magicbrush, zhao2024ultraedit} achieve controlled image manipulation by fine-tuning pretrained diffusion models on high-quality datasets, demonstrating impressive editing capabilities through specialized training.
Training-free methods offer a more flexible alternative, typically employing a two-stage process of latent inversion followed by denoising~\cite{dong2023prompt, hertz2022prompt, meng2021sdedit, tumanyan2023plug}. Recent work has focused on improving the accuracy of the inversion stage through advanced sampling techniques~\cite{avrahami2024stable, cao2023masactrl, li2023stylediffusion, tewel2024add, xu2024headrouter}. Notably, attention modulation~\cite{ju2023direct, lin2024schedule, miyake2025negative, mokady2023null, wang2024taming} has emerged as an effective tool in training-free methods due to its simplicity and precise control over editing operations. Building on these advances, our approach combines both paradigms: we first train a specialized model using a curated layer removal dataset, then manipulate the attention masks during inference to achieve accurate layer deletion.

\begin{figure*}[t]
    \centering
    \includegraphics[width=\textwidth]{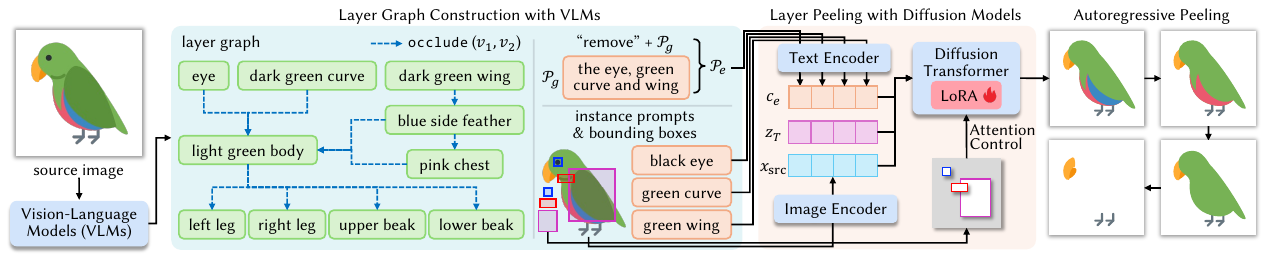}
    \caption{System diagram of \sysName. Given a rasterized source image (parrot), our system leverages vision-language models to analyze the image across three dimensions:
    (1) constructing a layer graph that captures occlusion relationships among color regions, (2) identifying the topmost layers and generating a global prompt $\mathcal{P}_g$ that describes them (then forming the editing prompt $\mathcal{P}_e$ by prepending the word ``\texttt{remove}'' to $\mathcal{P}_g$),
    and (3) decomposing the topmost layers into instance-level elements, each with an associated prompt and bounding box.
    The prompts and source image are encoded through text and image encoders to obtain $c_e$ and $x_\mathrm{src}$, respectively, while a noise vector $z_T$ is initialized.
    These inputs are fed into our LoRA-finetuned diffusion model, which synthesizes an image with the specified layers removed.
    The bounding boxes guide attention masks within joint attention modules.
    This layer removal process operates autoregressively, iteratively identifying and removing topmost layers until a complete layer-wise decomposition is achieved.
    The icon image is from \copyright SVGRepo.}
\label{fig:pipeline}
\end{figure*}

\section{\sysName}

In this section, we first introduce the definition and construction of a layer graph using VLMs to accurately detect non-occluded shapes (Section~\ref{sec:vlm}). We then describe the detailed architecture of our image diffusion model along with the localized attention sculpting technique for precise layer removal (Section~\ref{sec:image_diffusion}). Next, we present the autoregressive inference strategy and vectorization method (Section~\ref{sec:inference}).
Finally, we outline our dataset construction process, which enables diffusion models to perform the layer peeling task effectively (Section~\ref{sec:dataset}).

\subsection{Layer Graph Construction with VLMs}
\label{sec:vlm}

Although vision-language models excel at spatial understanding tasks, directly prompting them to detect non-occluded regions proves challenging (see the supplemental material).
To address this limitation, we introduce a \emph{layer graph} that explicitly encodes occlusion relationships among distinct color regions. This structured representation decomposes the top-layer detection task into sub-problems that VLMs can reliably solve, while providing an interpretable intermediate format that enables precise reasoning about spatial relationships between layers.

We define the layer graph as a directed graph $\mathcal{G} = (\mathcal{V}, \mathcal{E})$, where each vertex $v \in \mathcal{V}$ corresponds to an isolated, flat-color region within the image. 
Each vertex $v$ is associated with three attributes:
(1) Color -- a textual description that differentiates it as a separate element from adjacent regions;
(2) Label -- either a semantic description (\eg, "the bird's left eye") or a geometric identifier (\eg, "a red circle") when no clear semantic meaning is present;
and (3) \texttt{part\_of\_object(v,O)} -- a relational attribute indicating that region $v$ is a constituent of a larger semantic object $O$ (\eg, "a blue stripe on the hat" is part of the object "hat").

We define two types of relations (\ie, edges $\mathcal{E}$) between vertices in the layer graph:
\begin{itemize}
    \item \texttt{occlude($v_1, v_2$)}: A directed edge from vertex $v_1$ to $v_2$ indicates that the color region $v_1$ visually occludes -- either partially or fully -- the region $v_2$. This relation is fundamental for establishing the layering hierarchy and inferring depth order among regions.
    \item \texttt{interrupted\_shape($v_1, v_2$)}: This undirected, symmetric relation applies when two vertices $v_1$ and $v_2$ correspond to disconnected segments of a single, conceptually unified shape that has been visually separated by an intermediate occluding region. If the occluder were removed, $v_1$ and $v_2$ would form a continuous visual entity. For example, in Fig.~\ref{fig:pipeline}, the parrot's body is split into upper and lower parts by the dark green wing crossing it. This relation helps preserve semantic coherence by explicitly linking visually separated components that belong to the same object. 
    Although state-of-the-art VLMs can often infer such interrupted shapes as coherent whole entities without this explicit relation (as shown in Fig.~\ref{fig:vectorization}), we include it in our graph definition to ensure robust performance in complex occlusion scenarios.
\end{itemize}

Based on this graph definition, the VLM constructs $\mathcal{G}$ in several steps. First, it identifies all visible color regions in the source image, treating each as a vertex $v$. For each vertex, the VLM assigns the corresponding attributes: a color description, a label, and \texttt{part\_of\_object} attributes. Next, leveraging its spatial understanding capabilities, the VLM analyzes pairwise occlusion relationships between vertices to establish \texttt{occlude($v_1, v_2$)} and \texttt{interrupted\_shape($v_1, v_2$)} relations.
Finally, the VLM identifies non-occluded regions as vertices that are not targets of any \texttt{occlude($\cdot, v$)} edge. For these non-occluded regions, the VLM returns a concise global caption, denoted as $\mathcal{P}_g$.

While the initially constructed graph $\mathcal{G}$ captures the current occlusion relationships in the image, it must be dynamically updated throughout our proposed autoregressive peeling procedure, which progressively removes topmost layers to reveal those beneath.
Several factors necessitate ongoing graph maintenance: the emergence of previously occluded regions, potential VLM inaccuracies in the initial graph construction, and artifacts introduced by the image generation models. Specifically, after each editing operation alters the image, the VLM compares the existing graph $\mathcal{G}$ (from the previous iteration) with the current image.
It then verifies the validity of existing nodes and edges and makes corrections as needed.
This ensures that $\mathcal{G}$ remains an accurate and up-to-date representation of the image's evolving layered structure and occlusion relationships.

Furthermore, to facilitate localized attention control during inference (as detailed in Section~\ref{sec:image_diffusion}), the VLM is instructed to provide detailed information for each of the $N$ non-occluded regions identified in the global caption $\mathcal{P}_g$. For each such region $i$, this includes its 2D bounding box $\mathcal{B}_i$ and a corresponding instance-level caption label $\mathcal{P}_i$.
As a result, the VLM's output at each step consists of the tuple $\{\mathcal{G}, \mathcal{P}_g, \{\mathcal{B}_i, \mathcal{P}_i\}_{i=1}^N\}$, capturing the updated layer graph, a global description of the visible regions, and localized annotations for each non-occluded element, as illustrated in Fig.~\ref{fig:pipeline}.

\subsection{Layer Peeling with Diffusion Models}
\label{sec:image_diffusion}

As no existing pretrained image generation model supports the fine-grained image editing required for our task, \ie, the precise removal of individual layers, we fine-tune an image diffusion model to enable text-guided layer peeling.
We adopt the state-of-the-art Diffusion Transformer (DiT) architecture~\cite{peebles2023scalable}, specifically \texttt{FLUX.1-dev}, as our base model. This choice is motivated by its superior generation quality compared to previous models and its robust text-conditioning capabilities, enabled by its integration with the T5 text encoder~\cite{raffel2020exploring}.

During training, we sample triplets $(\mathcal{P}_e, I_\mathrm{src}, I_\mathrm{tar})$ from our custom-built dataset (Section~\ref{sec:dataset}). The editing prompt $\mathcal{P}_e$, source image $I_\mathrm{src}$, and target image $I_\mathrm{tar}$ are transformed into tokens using text and image encoders, yielding $c_e$, $x_\mathrm{src}$, and $x_\mathrm{tar}$, respectively.
For a given timestep $t\in [0,1]$, we sample a noise vector $\epsilon \sim \mathcal{N}(0, I)$ and compute the noised latent representation as $z=(1-t)x_\mathrm{tar}+t\epsilon$.
The noisy latent $z$ is then concatenated with the source token sequence $x_\mathrm{src}$ along the sequence dimension.
To encourage the model to learn spatial correspondences between the source and target images, we apply the \emph{same} positional encoding to both $z$ and $x_\mathrm{tar}$, following the approach of~\citet{huang2025photodoodle}.
We adopt the Conditional Flow Matching (CFM) loss from SD3~\cite{esser2024scaling} to optimize the model during training:
\begin{align}
    \mathcal{L}_\mathrm{CFM} = \mathbb{E}_{t, p_t(z | \epsilon), p(\epsilon)} || v_{\Theta}(z, x_\mathrm{src}, c_e, t) - u_t(z | \epsilon) ||_2^2,
    \label{eq:condflowmatch}
\end{align}
where $u_t(z | \epsilon)$ denotes the conditional vector field, and $v_{\Theta}(z, x_\mathrm{src}, c_e, t)$ represents the velocity field parameterized by the neural network's weights $\Theta$.
To avoid the computational cost of full-parameter finetuning, we employ Low-Rank Adaptation (LoRA)~\cite{hu2021lora}, a parameter-efficient training strategy.
LoRA freezes the pretrained model weights and injects trainable low-rank update matrices: $\Delta \mathbf{W} = \mathbf{A} \mathbf{B}$ into specified layers, where $\mathbf{A} \in \mathbb{R}^{m \times r}$, $\mathbf{B} \in \mathbb{R}^{r \times n}$, and the rank $r \ll \min(m,n)$. The final weight matrices are computed as $\mathbf{W}' = \mathbf{W} + \Delta \mathbf{W}$.

A key observation is that non-occluded regions in vector-style images often consist of numerous small, spatially disparate color areas. This granularity leads to highly fine-grained editing prompts, posing a significant challenge for the trained model, which must accurately associate each prompt instruction with its corresponding image region.
When this association fails, it results in \emph{erase failures} -- where layers remain visible in the output despite being marked for removal.
To mitigate this issue, we leverage instance-specific bounding boxes $\{\mathcal{B}_i\}_{i=1}^N$ and their associated labels $\{\mathcal{P}_i\}_{i=1}^N$ produced by the VLM (see Section~\ref{sec:vlm}) to modulate the model's attention mask during \textit{inference}.
Specifically, we concatenate the global editing prompt $\mathcal{P}_g$ with all instance-level labels $\{\mathcal{P}_g, \mathcal{P}_1, \dots, \mathcal{P}_N\}$ (see Fig.~\ref{fig:pipeline}).
These inputs are used alongside the bounding boxes to sculpt the joint attention map as follows:
\begin{itemize}
    \item \textbf{Global prompt $\mathcal{P}_g$ tokens} can attend to all other text tokens and the entire image, providing overarching guidance.
    \item \textbf{Instance label tokens $\mathcal{P}_i$} are constrained to attend only to image tokens within their associated bounding box $\mathcal{B}_i$ and to other tokens in $\mathcal{P}_i$. They cannot attend to unrelated instance labels $\mathcal{P}_j$ where $j \neq i$.
    \item \textbf{Image tokens} within a bounding box $\mathcal{B}_i$ must attend to their corresponding instance label tokens $\mathcal{P}_i$, while attention to labels $\mathcal{P}_j$ ($j \neq i$) is blocked. Attention among image tokens remains unrestricted to preserve global context and enable seamless recovery of occluded regions after layer removal.
\end{itemize}

The core insight is to establish a focused dialogue: each instance-specific instruction primarily "talks to" its designated image region, and vice versa, while minimizing interference from unrelated prompts or regions. This \emph{localized attention sculpting} technique effectively ensures that the model precisely identifies the specific regions to be removed, thereby improving the accuracy and fidelity of the layer peeling process.

\subsection{Inference}
\label{sec:inference}
Inference begins by sending the source image to VLMs to get the tuple $\{\mathcal{G}, \mathcal{P}_g, \{\mathcal{B}_i, \mathcal{P}_i\}_{i=1}^N\}$ and initializing a random noise vector $z_T$, which serves as the starting point for the denoising process.
The latents of the source image, denoted as $x_\mathrm{src}$, are kept noise-free during denoising steps to provide spatial correspondence and preserve high-frequency details.
Once the diffusion model completes the layer-peeling operation, it outputs an image in which the specified topmost layer has been removed.
To isolate the removed region, we compute a pixel-level difference map between the source and output images, apply thresholding ($\rho=20$ by default) to create a binary mask, and use this mask to extract the removed region from the source image.
The extracted region is then vectorized using an off-the-shelf vectorization tool.
This entire pipeline is executed autoregressively: after each layer is removed and vectorized, the resulting output image becomes the new source image for the next iteration. The process continues until the image becomes completely white.
Finally, we stack all the vectorized layers in reverse order to reconstruct the complete SVG.

\begin{figure}[t]
    \centering
    \includegraphics[width=\linewidth]{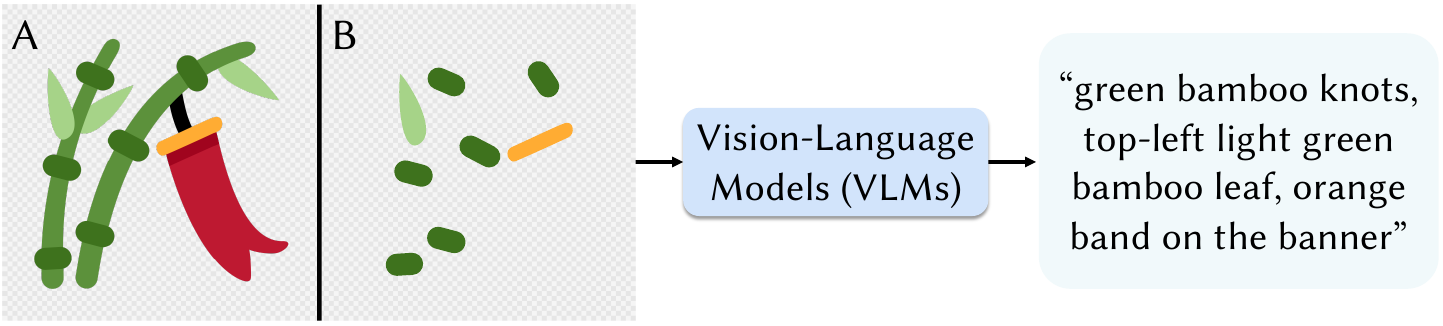}
    \caption{The visual annotation used for non-occluded layer captioning. The icon image is from \copyright SVGRepo.}
\label{fig:dataset}
\end{figure}

\subsection{\sysName~Dataset}
\label{sec:dataset}
To support effective training and future research on layer-by-layer peeling from vector-style images, we construct a new dataset: the \textbf{\sysName~Dataset}.
We focus on flat-color SVG, which feature clearly defined shape boundaries and structural regularity -- making them well-suited for validating our approach.

We curated a diverse collection of $179,000$ colored SVG from two major sources: SVGRepo ($52,800$ SVG) and Iconfont ($127,000$ SVG), encompassing a wide range of categories and stylistic variations.
Each SVG undergoes a multi-stage preprocessing pipeline:
(1) Grammar simplification to resolve incompatibilities with standard Python-based SVG-parsers, without altering visual appearance.
(2) Uniform resizing to a resolution of $512\times 512$ viewbox.
(3) Path filtering to discard SVG containing more than $30$ paths, as we observe these SVG typically contains excessive fine-grained details that hinder reliable detection and annotation.
Following these steps, we retain $115,700$ \emph{high-quality SVG}, which we split into training ($113,700$ SVG), validation ($1000$ SVG), and testing ($1000$ SVG) sets.

Our annotation process systematically identifies and extracts non-occluded paths within each simplified SVG, denoted as $S$.
We define that a path $p$ -- representing a shape -- is considered non-occluded if no other path both overlaps with it and appears above it in the layer hierarchy.
All such topmost paths are collected into a set $\{p_i\}_{i=1}^M$, where $M$ is the total number of non-occluded paths.
Both the complete SVG $S$ and the extracted layer group $\{p_i\}_{i=1}^M$ are rasterized using a renderer $\mathcal{R}(\cdot)$, yielding images $\mathcal{R}(S)$ and $\mathcal{R}(\{p_i\}_{i=1}^M)$, respectively.
To facilitate accurate VLM-assisted annotation, we design a visual prompting scheme (illustrated in Fig.~\ref{fig:dataset}).
Specifically, we arrange two images side by side: \texttt{A} (left), the SVG $\mathcal{R}(S)$, providing holistic context, and
\texttt{B} (right), the extracted non-occluded layers $\mathcal{R}(\{p_i\}_{i=1}^M)$, to be described.

The two panels are separated by a vertical black line to aid spatial parsing, and a checkerboard background is used to prevent shapes from blending with the background color.
We use \texttt{Gemini-2.0-Flash}, selected for its powerful vision-language performance and cost-efficiency, to generate two types of annotations:
semantic descriptions for recognizable elements in earlier stages and geometric descriptions for abstract shapes in later stages.
Each annotation is prefixed with ``\texttt{remove}'' to form an editing instruction $\mathcal{P}_e$.
For every step, we construct a training triplet $(\mathcal{P}_e, I_\mathrm{src}, I_\mathrm{tar})$, where $I_\mathrm{src}=\mathcal{R}(S)$ is the rasterized image before removal, $I_\mathrm{tar}=\mathcal{R}(S')$ represents the image after removal with $S'=S-\{p_i\}_{i=1}^M$, and $\mathcal{P}_e$ is the corresponding edit prompt.
This process iteratively repeats, removing and annotating topmost layers until the SVG is fully decomposed.
After a final cleaning phase, we obtain approximately $617,000$ high-quality data triplets, providing rich supervision for training models on fine-grained, layer-wise editing.

\section{Experiments}

\subsection{Baselines}
We compare \sysName~against six state-of-the-art optimization-based and learning-based methods for layer-wise image vectorization:
IVD~\cite{law2025image}, LIVE~\cite{ma2022towards}, O\&R~\cite{hirschorn2024optimize}, SGLIVE~\cite{zhou2024segmentation}, LIVSS~\cite{wang2024layered}, and LayerTracer~\cite{song2025layertracer}.
We exclude several other methods from our comparison due to practical limitations: some methods (\eg, Im2Vec~\cite{reddy2021im2vec}) only handle simple icons and fail on more complex inputs, while others lack a publicly available implementation~\cite{favreau2017photo2clipart, shen2021clipgen, chen2023editable, thamizharasan2024vecfusion}. Please refer to the supplementary material for implementation details and more comparisons (with~\citet{du2023image}).

\subsection{Evaluation Metrics}
We evaluate the quality of the generated vector graphics across three key dimensions:
\begin{itemize}
    \item \textit{Path Semantics}. To evaluate the semantic importance of individual shapes, we first obtain captions for each test image using a VLM.
    We then randomly remove $30\%$ of paths from the SVG~\cite{zhang2024text} and measure the drop in CLIP text-image similarity scores~\cite{radford2021learning} between the original and modified renderings.
    A larger score drop indicates that individual shapes contribute more significantly to the overall semantic meaning.
    \item \textit{Path Irregularity}. Using the ground truth SVG as a reference, we compute the Chamfer Distance (CD) between each path in the generated SVG and its closest counterpart in the ground truth SVG (based on the minimal CD). The average CD across all paths quantifies path distortion -- higher values indicate greater deviation from the ground truth paths.
    \item \textit{Visual Fidelity}. We evaluate how closely the generated SVG matches the source image by computing both Mean Squared Error (MSE) and LPIPS~\cite{zhang2018unreasonable} (VGG backbone~\cite{simonyan2014very}) between the rendered vector graphics and the original image.
\end{itemize}

\begin{table}[t]
    \centering
    \scriptsize
    \caption{Quantitative comparison of layer-wise image vectorization methods across path semantics, irregularity, and visual fidelity (\ie, MSE and LPIPS).}
    \begin{tabular}{lcccc}
        \toprule
        Method & 
        \begin{tabular}[c]{@{}c@{}}Path\\Semantics\end{tabular} $\big\uparrow$ &
        \begin{tabular}[c]{@{}c@{}}Path\\Irregularity\end{tabular} $\big\downarrow$ &
        MSE $\big\downarrow$ & 
        LPIPS $\big\downarrow$ \\
        \toprule
        IVD~\cite{law2025image} & $0.0183$ & $69.17$ & $0.0135$ & $0.0610$\\
        LIVE~\cite{ma2022towards} & $0.0158$ & $96.41$ & $0.0033$ & $0.0388$\\
        O\&R~\cite{hirschorn2024optimize} & $0.0196$ & $149.9$ & $0.0095$ & $0.0694$\\
        SGLIVE~\cite{zhou2024segmentation} & $0.0172$ & $99.19$ & $0.0034$ & $0.0415$\\
        LIVSS~\cite{wang2024layered} & $0.0075$ & $80.58$ & $\mathbf{0.0006}$ & $0.0099$\\
        LayerTracer~\cite{song2025layertracer} & $0.0121$ & $74.23$ & $0.0508$ & $0.0822$\\
        \hline
        \sysName~(Ours) & $\mathbf{0.0242}$ & $\mathbf{25.41}$ & $0.0011$ & $\mathbf{0.0083}$\\
        \bottomrule
    \end{tabular}
    \label{tab:comparison}
\end{table}

\subsection{Comparison with Existing Methods}
\subsubsection{Quantitative Comparison}
Table~\ref{tab:comparison} presents a comprehensive evaluation of all baseline methods. Our \sysName~significantly outperforms existing approaches in path semantics and path regularity, reflecting its ability to preserve meaningful shape information and reconstruct smooth, coherent paths.
These improvements are primarily attributed to our autoregressive peeling paradigm, which incrementally recovers complete shapes while preserving the logical order of visual layers.
In terms of visual fidelity, our method achieves competitive performance relative to the optimization-based method, LIVSS, that directly minimize MSE. This is made possible by our use of shared positional encodings and localized attention control, which help the model maintain fine-grained alignment with the original image.
Overall, the results demonstrate that \sysName~achieves strong semantic understanding and visual quality, producing vector graphics that are both structurally coherent and visually faithful to the source image.

\subsubsection{Qualitative Comparison}
Fig.~\ref{fig:comparison} presents a side-by-side visual comparison of the outputs from various baseline methods, our \sysName, and the ground truth SVG.
To highlight structural differences, we render a black outline along the contour of each shape.
(1) The optimization-based geometric method, IVD, lacks semantic awareness in its curvature-guided inpainting process, often leading to unnatural reconstructions (\eg, the distorted and incomplete angel wings in the first column of Fig.~\ref{fig:comparison}).
(2) For optimization-based methods in the differentiable rendering category (LIVE, O\&R, SGLIVE), a common strategy is to initialize shapes at the centers of mass of poorly reconstructed regions. However, these methods often lead to excessive use of shapes to approximate the source image, fragmenting semantically meaningful elements, as evident in all four examples.
LIVSS relies on the feature-averaging effect of SDS loss~\cite{poole2022dreamfusion} to remove high-frequency details and derive abstracted versions of images.
It then uses segmentation models to extract the initial SVG. However, this abstraction strategy is less effective on flat-color, highly abstract SVG, resulting in incomplete (\eg, the cake top) or disconnected shapes (\eg, handle of the knife) that deviate from human perception.
(3) The learning-based method LayerTracer generates a $3 \times 3$ image grid to depict a progressive layer construction sequence.
However, it does not explicitly model occlusion relationships, which can lead to inconsistent layering -- multiple occluded shapes may appear between successive frames. Furthermore, the lack of pixel alignment across frames introduces jagged boundaries.

In contrast, our method leverages VLM reasoning to determine the appropriate number of layers and uses autoregressive peeling with semantic guidance to reconstruct occluded regions, resulting in cleaner, semantically coherent layers.
Notably, our method achieves strong structural similarity compared to the ground truth SVG, while occasionally offering novel interpretations of shapes, as demonstrated by the regularized structure of the stamen in the sakura example.

\begin{figure}[t]
    \centering
    \includegraphics[width=0.9\linewidth]{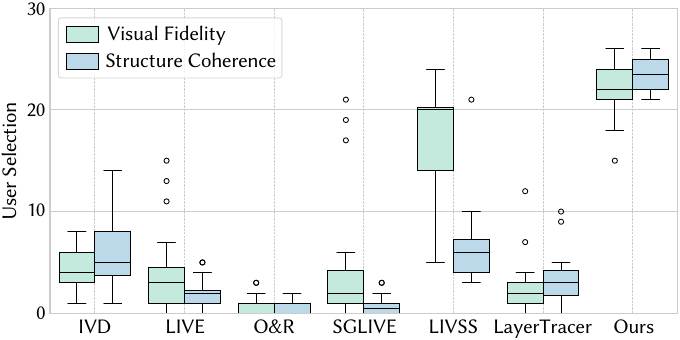}
    \caption{Perceptual study results. Our method achieves the highest number of user selections across both visual fidelity and structural coherence.}
\label{fig:user_study}
\end{figure}

\subsubsection{Perceptual Study}
To quantitatively evaluate the improvements in layer-wise image vectorization quality achieved by \sysName, we conducted a formal subjective user study focusing on two aspects:
(1) visual fidelity and (2) structural coherence.
We randomly selected $20$ images from the test set and vectorized them using six baseline methods and \sysName.
In the first task, we presented participants with the source image and all vectorized outputs, and asked them to select all options that preserved the original visual appearance.
In the second task, we rendered videos illustrating the progressive construction of each SVG (see supplemental material for details) and asked participants to identify all options that demonstrated logically layered structures.
The study was conducted via online questionnaires and received feedback from $30$ participants ($17$ male, $13$ female, average age $27.73$).
Notably, $46.67\%$ of participants reported prior experience in graphic design.
To reduce fatigue, participants were allowed to complete the study at their own pace, with optional breaks.
Fig.~\ref{fig:user_study} presents the results using box plots to show the distribution of user selections across all $20$ examples.
\sysName~has a higher selection ratio than all baselines in both tasks, indicating that it more effectively preserves visual fidelity and produces structurally coherent SVG.

\begin{table}[t]
    \centering
    \footnotesize
    \caption{Quantitative results of the ablation study.}
    \begin{tabular}{lcccc}
        \toprule
        Method & 
        \begin{tabular}[c]{@{}c@{}}Path\\Semantics\end{tabular} $\big\uparrow$ &
        \begin{tabular}[c]{@{}c@{}}Path\\Irregularity\end{tabular} $\big\downarrow$ &
        MSE $\big\downarrow$ & 
        LPIPS $\big\downarrow$ \\
        \toprule
        \texttt{Claude-Sonnet-4.0} & $0.0232$ & $26.65$ & $0.0014$ & $0.0096$\\
        \texttt{Qwen2.5-VL-32B-Instruct} & $0.0219$ & $30.56$ & $0.0023$ & $0.0096$\\
        w/o Layer Graph & $0.0218$ & $37.41$ & $0.0133$ & $0.0261$\\
        w/o Attention Control & $0.0209$ & $38.43$ & $0.0126$ & $0.0481$\\
        \hline
        Default & $\mathbf{0.0242}$ & $\mathbf{25.41}$ & $\mathbf{0.0011}$ & $\mathbf{0.0083}$\\
        \bottomrule
    \end{tabular}
    \label{tab:ablation}
\end{table}

\begin{figure}[t]
    \centering
    \includegraphics[width=0.9\linewidth]{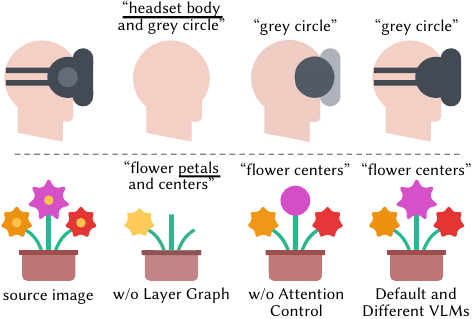}
    \caption{Qualitative results of the ablation study. We show the editing prompts above each image and the editing results. The icons are from \copyright SVGRepo.
    }
\label{fig:ablation}
\end{figure}

\subsection{Ablation Study}
We perform ablation experiments to evaluate the effectiveness of key components in our pipeline.
We first replace the default VLM, \texttt{Gemini-Flash-2.5}, with two alternatives: \texttt{Claude-Sonnet-4.0} and \texttt{Qwen-2.5-VL-32B-Instruct}, and report the quantitative results in Table~\ref{tab:ablation}. The results show that our method remains robust across different VLM choices. Moreover, since \texttt{Qwen} is an open-source model, it provides a practical alternative to proprietary VLMs.
We then examine the importance of the layer graph by removing this component and instead prompting VLMs to directly analyze (check the supplemental material) the set of non-occluded layers. Without the guidance of a layer graph, VLMs often misidentify partially occluded layers as top layers (see Fig.~\ref{fig:ablation}), resulting in incorrect layer ordering. Since our diffusion model is trained specifically to operate in non-occluded regions, these misidentifications lead to a degraded output quality, evident in the drop in path regularity and visual fidelity scores in Table~\ref{tab:ablation}.
Next, we assess the impact of attention-guided control. Disabling this component results in two recurring failure modes: (1) incomplete removal of the intended layer and (2) unintended modifications to unrelated regions (\eg, the purple flower in Fig.~\ref{fig:ablation}). Both issues lead to performance degradation across all evaluation metrics.
Notably, path semantics remain relatively consistent across all ablations. We attribute this inherent robustness to the design of our progressive peeling pipeline, which enforces a clear and interpretable correspondence between shapes and prompts throughout the process.

\subsection{Evaluation of Generalizability}
We provide examples of the autoregressive peeling process for images with different layer complexities in Fig.~\ref{fig:more_peeling}.
Additionally, our system demonstrates robust performance and generalizability across a diverse range of SVG styles in Fig.~\ref{fig:more_style}. The source images in this subsection are downloaded from the Flaticon website, or are from the test set, and thus are not in the training set.

\begin{figure}[t]
    \centering
    \includegraphics[width=1.0\linewidth]{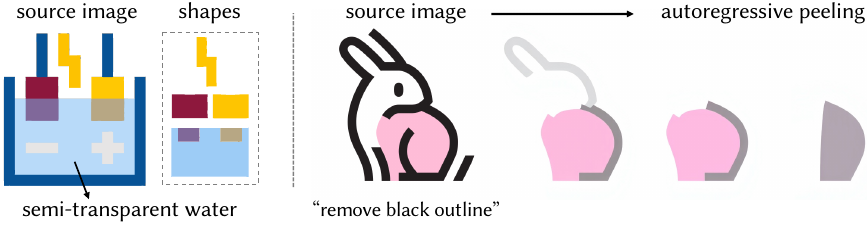}
    \caption{Limitations of our method include the fragmentation of semi-transparent layers into multiple discrete shapes (left panel) and error accumulation during the autoregressive process (right panel), which can produce redundant paths in the final SVG output. The icons are from \copyright SVGRepo.}
\label{fig:limitation}
\end{figure}

\section{Conclusion and Discussion}
In this work, we have introduced \sysName, a novel framework for layer-wise image vectorization.
\sysName~first uses VLMs to construct a layer graph on the input image, enabling accurate captioning of non-occluded regions.
It then employs a finetuned image diffusion model to remove the identified layers and faithfully reconstruct the underlying occluded regions.
To enable precise and robust editing, we propose a localized attention control mechanism that enhances the fidelity of the removal process.
We also curated a new dataset to support future research on autoregressive layer peeling.
Extensive experiments confirm the effectiveness of \sysName~in producing high-quality, structurally coherent vector graphics.

Since our collected dataset only contains flat-color SVG, our system's performance may degrade when handling semi-transparent elements (see Fig.~\ref{fig:limitation}).
Additionally,
there are stochastic errors (see supplemental materials for details) in the system arising from incorrect top-layer detection by VLMs and the diffusion model’s inability to accurately remove specified regions, which leads to error accumulation in the autoregressive pipeline.
Our method may also struggle with realistic images, primarily because no ground-truth layer-peeling dataset exists for photographs, and VLMs often have difficulty accurately identifying layers in natural images.
In the future, we aim to apply data augmentation techniques, such as adding gradient colors, to improve the model's generalizability.
To mitigate error accumulation, we propose implementing an automatic success check after each round of peeling for validation.

\begin{acks}
The work described in this paper was fully supported by a GRF grant from the Research Grants Council (RGC) of the Hong Kong Special Administrative Region, China [Project No. CityU 11216122].
\end{acks}

\bibliographystyle{ACM-Reference-Format}
\bibliography{ref}

\input{figure_page}
\clearpage

\appendix

\input{appendix}

\end{sloppypar}

\end{document}

%% file: figure_page.tex
\begin{figure*}[t]
    \centering
    \includegraphics[width=0.95\textwidth]{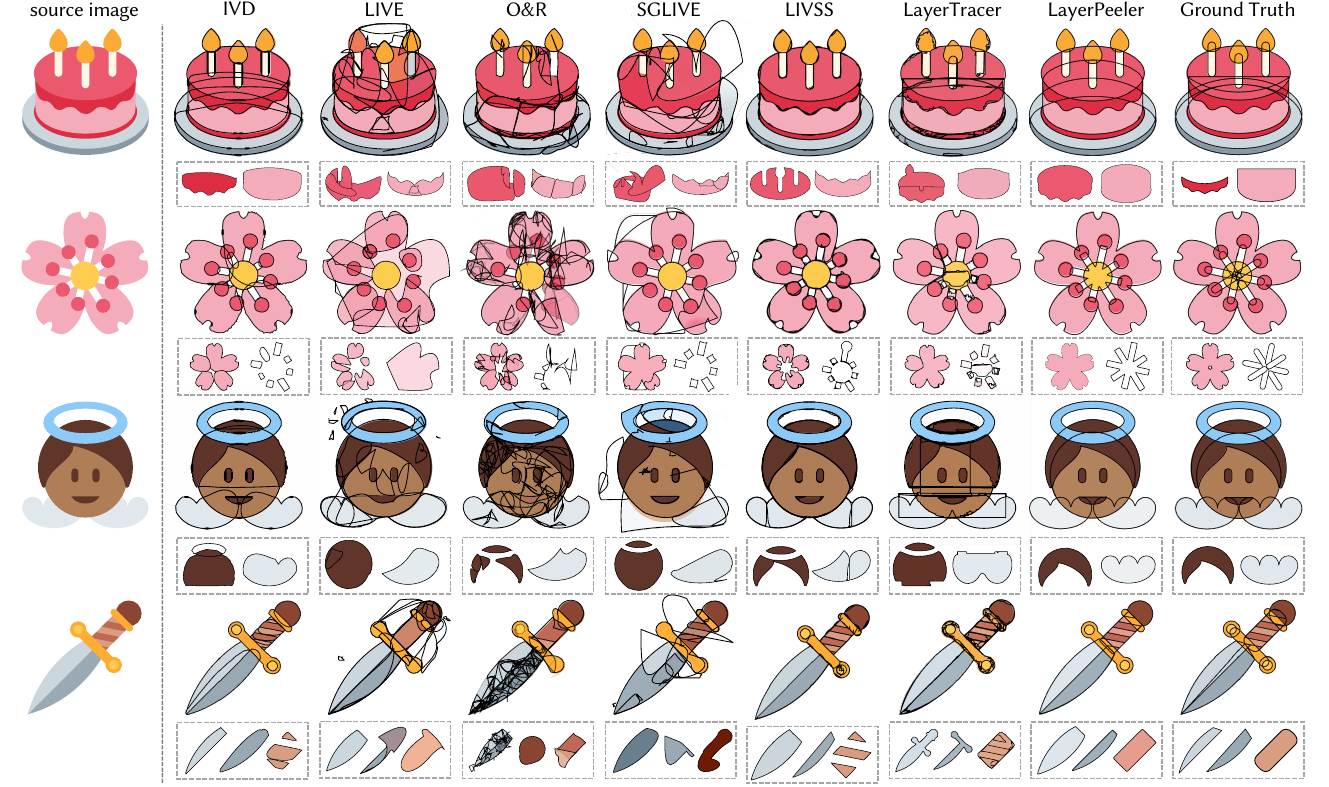}
    \caption{Qualitative comparison between baselines and \sysName. We render black strokes along the contours of recovered vector elements to better illustrate their structural quality.
    The rule-based method IVD lacks semantic understanding of each shape; thus, its curvature-based inpainting leads to unnatural completions like the wing of the angel.
    Training-free methods (particularly LIVE, O\&R, and SGLIVE) tend to produce fragmented shapes with irregular components due to their naive depth estimation, as evident in the flower petals.
    LIVSS's blurring strategy does not work for non-photorealistic icon images, yielding disjoint or incomplete shapes (\eg, the head of the angel).
    LayerTracer trains a model to predict the SVG construction sequence, but the alignment between two frames is not guaranteed.
    In contrast, \sysName~generates paths that are both semantically meaningful and geometrically regular.
    }
\label{fig:comparison}
\end{figure*}

\begin{figure*}[t]
    \centering
    \includegraphics[width=0.95\textwidth]{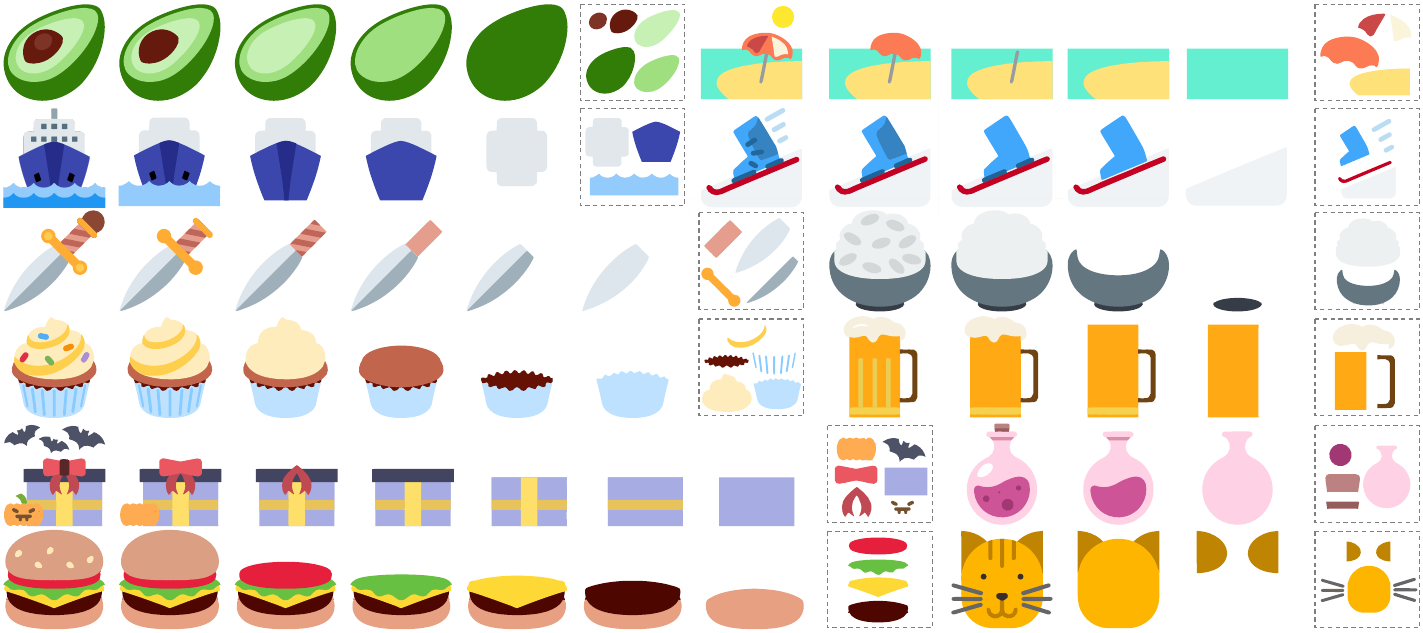}
    \caption{Additional examples demonstrating our autoregressive peeling process on images with varying layer complexity.}
\label{fig:more_peeling}
\end{figure*}

\begin{figure*}[t]
    \centering
    \includegraphics[width=\textwidth]{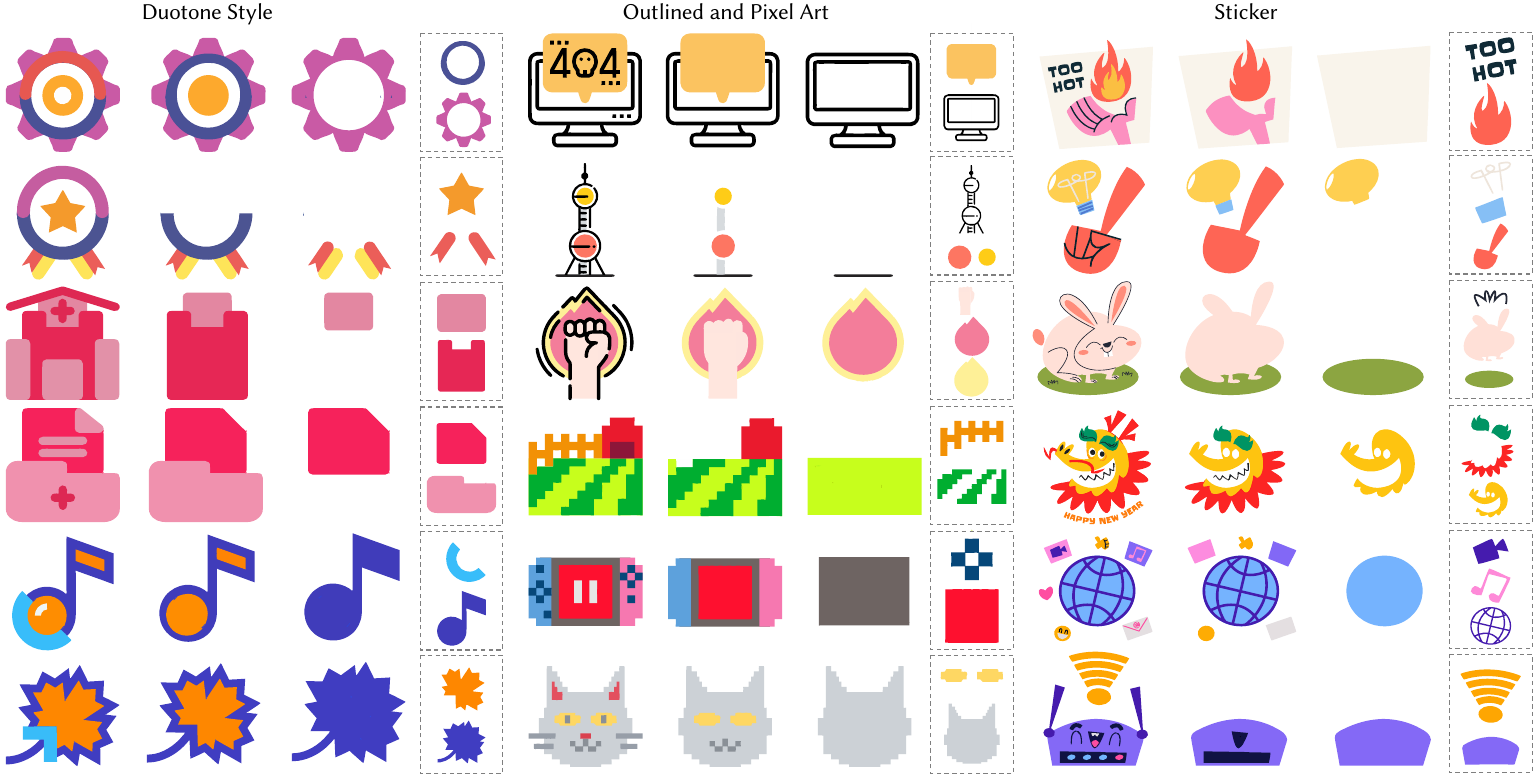}
    \caption{Evaluation of generalizability across different styles of source images, including flat-color icons, pixel art, and stickers.
    }
\label{fig:more_style}
\end{figure*}

\begin{figure*}[t]
    \centering
    \includegraphics[width=\textwidth]{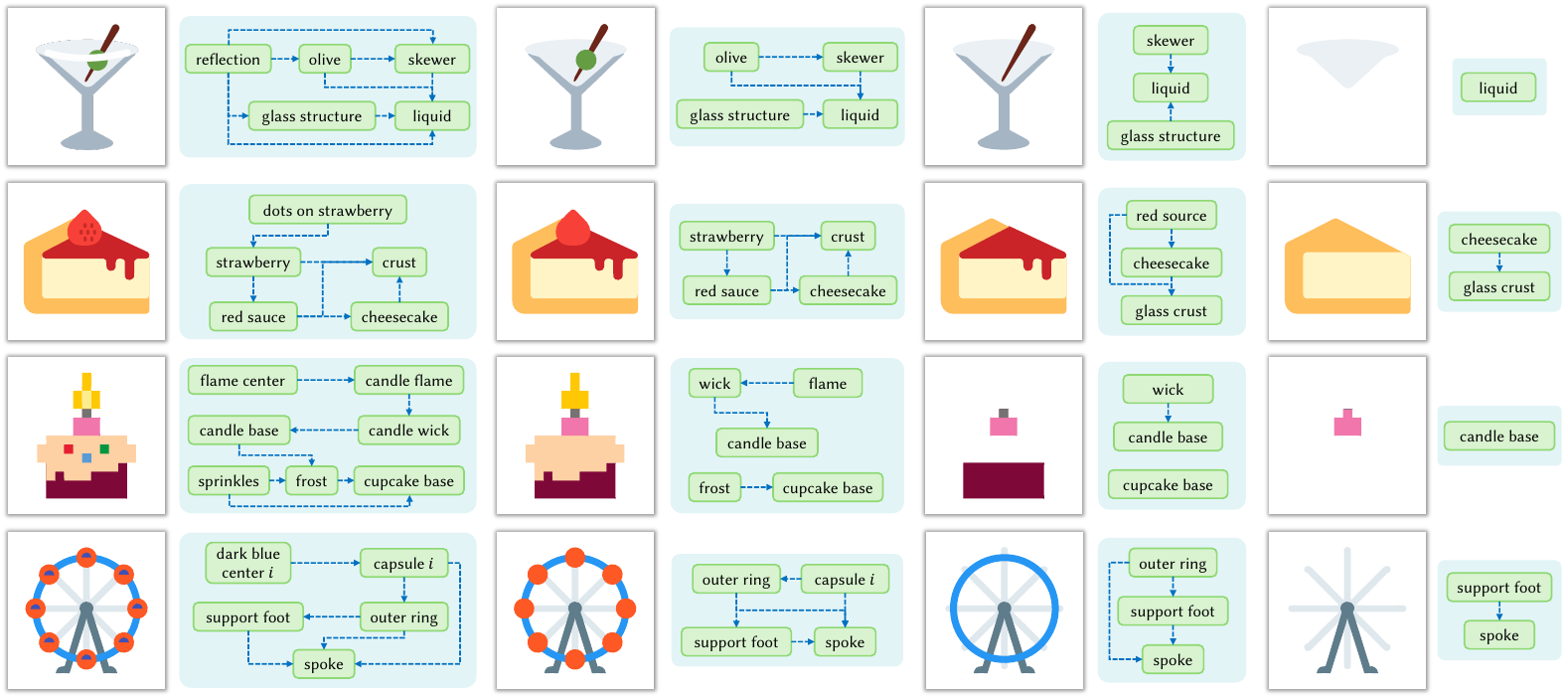}
    \caption{Additional examples demonstrating the construction and progressive updates of layer graphs throughout the peeling process.}
\label{fig:more_layer_graph}
\end{figure*}

%% file: appendix.tex
\definecolor{codegreen}{rgb}{0,0.6,0}
\definecolor{codegray}{rgb}{0.5,0.5,0.5}
\definecolor{codepurple}{rgb}{0.58,0,0.82}
\definecolor{backcolour}{rgb}{0.961,0.984,0.984}

\lstdefinestyle{mystyle}{
  backgroundcolor=\color{backcolour},
  commentstyle=\color{codegreen},
  keywordstyle=\color{magenta},
  stringstyle=\color{codepurple},
  basicstyle=\ttfamily\footnotesize,
  breakatwhitespace=false,
  breaklines=true,
  captionpos=b,
  keepspaces=true,
  numbers=none, 
  showspaces=false,
  showstringspaces=false,
  showtabs=false,
  tabsize=2,
  frame=ltb,
  framerule=0pt,
}

\lstset{style=mystyle}

\section{Overview}
This appendix contains supplementary implementation details and experimental results.

\section{Implementation Details}
For training, we initialize the FLUX weights from OmniEditor~\cite{huang2025photodoodle}, which was pretrained on the SeedEdit~\cite{ge2024seed} dataset to provide basic editing capabilities for natural images.
We adopt LoRA with a rank of $256$ and a learning rate of $0.0001$.
We use learning rate warmup followed by cosine decay and train with a minibatch size of $32$.
Training runs for $50,000$ steps on four NVIDIA A100 GPUs using our custom \sysName~dataset.
During inference, we employ \texttt{Gemini-2.5-Flash} for constructing layer graphs.
For image generation, we use $40$ diffusion steps with a guidance scale of $4.5$.
For vectorization, we use the tool provided by Recraft\footnote{\url{https://www.recraft.ai/}}, selected for its high-quality output and seamless code integration. However, we emphasize that our approach is agnostic to the specific vectorization method used.
In our experiments, we use a test set comprising $250$ randomly selected images rendered from SVG, ensuring no overlap with the training dataset.

\section{Runtime Analysis}
On an NVIDIA RTX $5880$ Ada GPU, the average vectorization time per image is $128.05$ seconds. Each peeling iteration takes approximately $26.2$s for the VLM, $15$s for the diffusion model, and $4.7$s for vectorization. Notably, the number of iterations required for complex images is often comparable to that for simpler ones: images with more paths may have similar (or even lower) layer complexity than those with fewer paths. As a result, the overall runtime remains relatively stable. Our pipeline is thus faster than optimization-based approaches, which typically require over $10$ minutes per image.

\section{More Comparisons}
\citet{du2023image} proposed an optimization-based geometric method that takes a user-provided segmentation map as input and generates an SVG with gradient colors and semi-transparent layers. To recover occluded regions, the method assumes that upper layers are semi-transparent. Consequently, it is limited to simple overlap scenarios, depends heavily on per-pixel segmentation masks, and lacks a dedicated shape-completion step. In other words, while it can merge visible regions in the segmentation map to form a complete shape, it cannot reconstruct fully invisible regions, since these never appear in the segmentation maps.

\begin{figure}[t]
    \centering
    \includegraphics[width=0.9\linewidth]{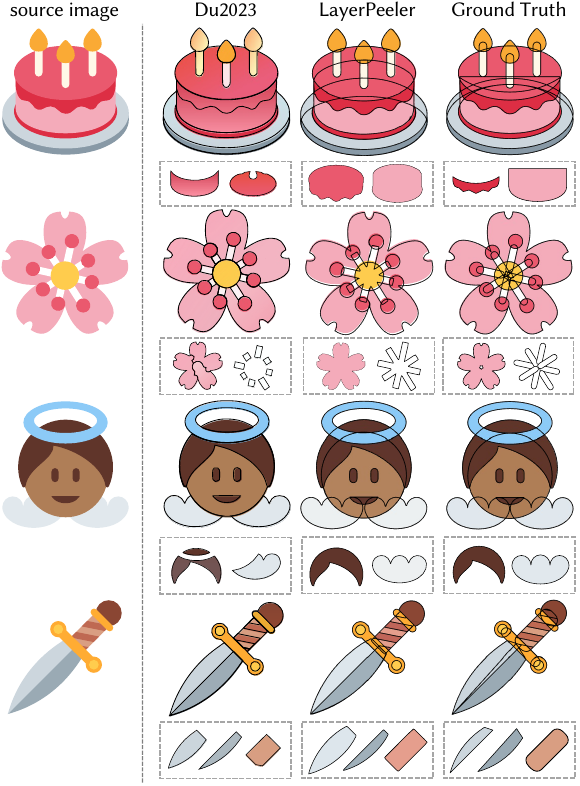}
    \caption{Qualitative Comparison with~\citet{du2023image}. At first glance, their method appears to produce clean outlines, but this actually reflects incomplete shapes (\ie, no reconstruction occurs in invisible regions that are not indicated by the segmentation map). The icons are from \copyright SVGRepo.
    }
\label{fig:more_comparison}
\end{figure}

When comparing \sysName~with their method, we encountered a known issue in the released code\footnote{\url{https://github.com/Zhengjun-Du/ImageVectorViaLayerDecomposition/issues/1}} that prevents the generation of the final SVG output. As this issue remains unresolved, only $195$ out of $250$ input images successfully produced SVG results. Therefore, we omit a direct comparison in the main paper and instead provide a side-by-side comparison in Fig.~\ref{fig:more_comparison}.
The results show that while their method benefits from segmentation maps to produce clean shape boundaries, the absence of a true shape completion step leads to incomplete reconstructions (\eg, the top of the cake and the angel’s wing). In practice, their method can only handle simple overlaps (\eg, the blade of the dagger), whereas our method explicitly targets the reconstruction of invisible regions.

We also report quantitative results of~\citet{du2023image} on the $195$ test images in Table~\ref{tab:more_comparison}. The Path Irregularity score indicates that their method fails to generate shapes consistent with the ground-truth SVG due to the limitation of their method.

\begin{table}[t]
    \centering
    \scriptsize
    \caption{Quantitative comparison with~\citet{du2023image}.}
    \begin{tabular}{lcccc}
        \toprule
        Method & 
        \begin{tabular}[c]{@{}c@{}}Path\\Semantics\end{tabular} $\big\uparrow$ &
        \begin{tabular}[c]{@{}c@{}}Path\\Irregularity\end{tabular} $\big\downarrow$ &
        MSE $\big\downarrow$ & 
        LPIPS $\big\downarrow$ \\
        \toprule
        \citet{du2023image} & $0.0216$ & $44.08$ & $0.0043$ & $0.0173$\\
        \sysName~(Ours) & $\mathbf{0.0242}$ & $\mathbf{25.41}$ & $\mathbf{0.0011}$ & $\mathbf{0.0083}$\\
        \bottomrule
    \end{tabular}
    \label{tab:more_comparison}
\end{table}

\section{Complex Images}
\begin{figure*}[t]
  \centering
  \includegraphics[width=0.95\textwidth]{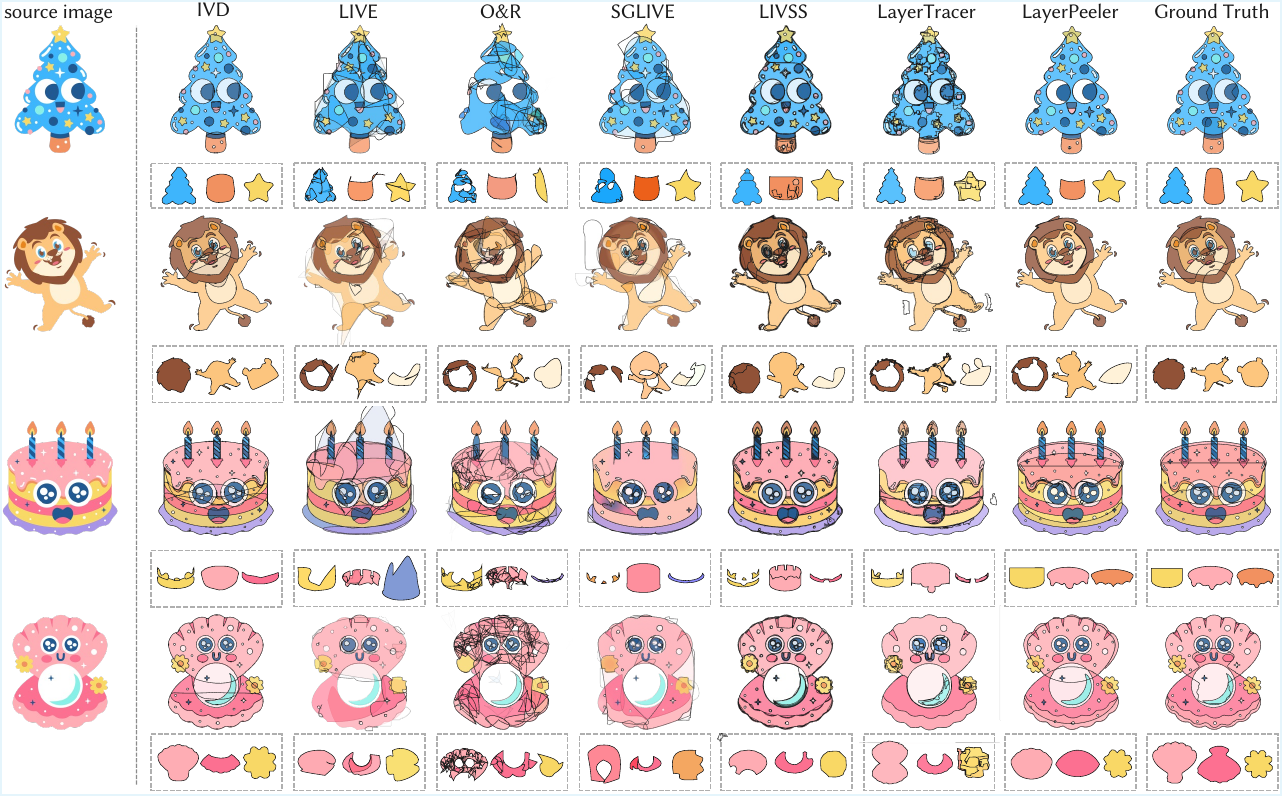}
  \caption{
    Qualitative comparison of complex images. These results demonstrate that LayerPeeler generalizes well to more complex stickers. The sticker images are from \copyright Flaticon.
  }
\label{fig:comparison_complex}
\end{figure*}

Our primary test set consists mainly of flat-color icons. To provide a fairer evaluation, we additionally collect a set of $20$ complex stickers (averaging $46$ paths, compared to $10.8$ in the original test set) and evaluate all baselines on them. As image complexity increases, LayerPeeler consistently outperforms competing methods (see Table~\ref{tab:comparison_complex}).
In contrast, optimization-based approaches (\eg, LIVE, O\&R, and SGLIVE) can process complex images by introducing a large number of paths, but their outputs often suffer from fragmented shapes and disorganized layer structures, as reflected in the Path Semantics and Path Irregularity metrics. Moreover, Fig.~\ref{fig:comparison_complex} shows that the visual trends observed with simple icons persist in these more complex cases.
Overall, these results demonstrate LayerPeeler’s ability to preserve coherent structure and semantic organization even for highly complex inputs. Notably, we exclude~\citet{du2023image} from this comparison, since its runtime grows \textit{exponentially} with the number of regions, preventing it from producing results within a reasonable time.

\begin{table}[h]
  \centering
  \scriptsize
  \caption{Quantitative comparison over complex images.}
  \begin{tabular}{lcccc}
      \toprule
      Method & 
      \begin{tabular}[c]{@{}c@{}}Path\\Semantics\end{tabular} $\big\uparrow$ &
      \begin{tabular}[c]{@{}c@{}}Path\\Irregularity\end{tabular} $\big\downarrow$ &
      MSE $\big\downarrow$ & 
      LPIPS $\big\downarrow$ \\
      \toprule
      IVD~\cite{law2025image} & $0.0194$ & $28.97$ & $0.0083$ & $0.0319$\\
      LIVE~\cite{ma2022towards} & $0.0191$ & $42.18$ & $0.0074$ & $0.0633$\\
      O\&R~\cite{hirschorn2024optimize} & $0.0198$ & $46.14$ & $0.0229$ & $0.1214$\\
      SGLIVE~\cite{zhou2024segmentation} & $0.0187$ & $42.33$ & $0.0088$ & $0.0622$\\
      LIVSS~\cite{wang2024layered} & $0.0099$ & $30.71$ & $\mathbf{0.0014}$ & $0.0270$\\
      LayerTracer~\cite{song2025layertracer} & $0.0118$ & $36.11$ & $0.0426$ & $0.1121$\\
      \hline
      \sysName~(Ours) & $\mathbf{0.0244}$ & $\mathbf{9.733}$ & $0.0026$ & $\mathbf{0.0149}$\\
      \bottomrule
  \end{tabular}
  \label{tab:comparison_complex}
\end{table}

\section{Stochastic Errors}
Our \sysName~is built upon powerful VLMs and image diffusion models, both of which can introduce stochastic errors during execution.
To quantify these errors, we sample $30$ icon images from the test set and vectorize each image three times using different random seeds. Across a total of $318$ peeling iterations, we observe $0$ invalid JSON errors, $8$ incorrect top-layer detection errors, and $19$ diffusion model errors.
These results indicate that the diffusion process is the primary bottleneck of our pipeline.
To alleviate stochastic errors, our method already employs several strategies, including layer graph representation and updating, local/global attention control, and a positional encoding scheme. Furthermore, our approach is not restricted to a specific model version: the vectorization paradigm can be adapted to more advanced models (\eg, the recently introduced \texttt{Flux.1-Kontext}), potentially yielding more robust performance and lower error rates.

\section{More Results}
\label{sec:more_results}
For better accessibility and visualization, we have created a static project website in the supplementary material. The website contains:
\begin{itemize}
  \item Complete vectorization results on the test set of $250$ icons;
  \item Side-by-side comparisons with baseline methods;
  \item Detailed user study results.
\end{itemize}

\section{System Prompts}
\label{sec:system_prompt}

This section lists the system prompts used by the \sysName~system for various Vision-Language Model tasks, covering initial layer graph construction (Section\ref{subsection:layer_graph_construction}), layer graph maintenance (Section~\ref{subsection:layer_graph_update}), instance prompt and bounding box generation (Section~\ref{subsection:box_detection}), isolated layer annotation (Section~\ref{subsection:layer_annotation}), and the specific prompt for the ablation study (Section~\ref{subsection:ablation_prompt}).

\subsection{Initial Layer Graph Construction}
\label{subsection:layer_graph_construction}
This system prompt guides the VLM to analyze a flat-color cartoon image, construct a layer graph representing distinct color regions and their spatial/occlusion relationships, identify non-occluded elements based on this graph, and generate a corresponding caption.

\begin{lstlisting}
You are an advanced Vision-Language Model tasked with analyzing cartoon-style, flat-color images to construct a simplified 2D layer graph and identify non-occluded visual elements. Your analysis will form the basis for a layered image decomposition process, where layers correspond to distinct color regions.

**Core Task:**
1.  **Image Analysis:** Analyze the provided flat-color, cartoon-style image.
2.  **Layer Graph Construction:** Infer and describe a 2D layer graph.
    * **Nodes:** Identify distinct, continuous flat-color regions within the image. **Each such visually separable color region constitutes a node.** These regions are the fundamental building blocks for the layered structure.
    * **Node Attributes:** For each node (color region), describe:
        * Its dominant color (e.g., "red," "light blue").
        * A semantic label if the region clearly corresponds to a recognizable object part (e.g., "the character's skin," "the hat's brim," "the left eye's pupil," "a highlight on the sphere").
        * If no clear semantic label applies, use a geometric description (e.g., "large circular red area," "thin black outline").
        * `part_of_object(N, O)`: Color-region Node N is recognized as a component of a larger semantic object O (e.g., a 'blue hat body' `part_of_object` 'hat'; a 'highlight in the eye' `part_of_object` 'eye'). This helps group related color regions conceptually.
    * **Relationships (Edges):** Define the spatial and occlusion relationships between these color-region nodes. Key relationship types are:
        * `occludes(A, B)`: Color-region Node A visually covers (partially or fully) Color-region Node B.
        * `interrupted_shape(A, B)`: Color-region Node A and Color-region Node B share the same color and are conceptually part of the same larger shape, but are visually disconnected because another region occludes the space and splits the shape into two or more parts. If the occluding region(s) were removed, A and B would form a single continuous visual entity. If A and B are interrupted shapes, then it means there is another color region that occludes the space between them.
3.  **Non-Occluded Element Identification:** Based *solely* on the `occludes` relationships in your constructed layer graph, identify all color-region nodes that are not occluded by any other node.
4.  **Caption Generation:** Provide a concise caption describing these non-occluded color-region nodes, using their semantic or geometric descriptions.

**Instructions & Definitions:**

* **Node Granularity:** **Each continuous patch of flat color must be treated as a distinct node.** If an apparent 'object' (e.g., a 'hat') is composed of multiple color areas (e.g., 'blue hat body,' 'yellow hat band,' 'white highlight on hat'), each of these color areas must be a separate node in the graph. Do not merge distinct color regions into a single 'object' node at the graph construction stage.
* **Non-Occluded Element Definition (Graph-based):** A color-region node in the layer graph is considered non-occluded if and only if there are no `occludes(X, Node)` relationships where `Node` is the element in question (i.e., no other color-region node X occludes it).
* **Background Exclusion:** The global white background of the canvas is not considered a color-region node for the layer graph.
* **Internal Details:** White or other colored shapes *inside* what might be perceived as a larger object (e.g., highlights, pupils within eyes, patterns on clothes) are themselves distinct color-region nodes. Their relationship to surrounding/underlying color-region nodes must be captured.
* **Concise Captioning:**
    * If multiple, similar non-occluded color-region nodes exist and belong to the same semantic object (e.g., "the white highlight on the left eye," "the white highlight on the right eye"), group them in the caption (e.g., "the white highlights on the eyes").
    * Describe the elements using their semantic labels or geometric descriptions. **Avoid using the word 'region' in the final caption** (e.g., "the blue hat band," not "the blue hat band region").
    * The caption should be LESS than 40 words.

**Output Format:**

* **Image Description:** `<image_description>...</image_description>`
    Provide a brief overall description of the image content.

* **Layer Graph Construction Reasoning:** `<layer_graph_reasoning>...</layer_graph_reasoning>`
    Describe your thought process for segmenting the image into flat color-region nodes and the visual cues used to determine their relationships, especially occlusion. Explain how you decided on the chosen nodes and edges.

* **Layer Graph (Simplified JSON-like Representation):** `<layer_graph>...</layer_graph>`
    Represent the graph with nodes and edges. Use unique IDs for nodes.
    Example (reflecting color-region nodes for a hat with a band and star):
    ```json
    {
      "nodes": [
        {"id": "N1", "description": "character's head skin", "color": "peach"},
        {"id": "N2", "description": "upper hat body", "color": "blue", "part_of_object": "hat"},
        {"id": "N3", "description": "hat band", "color": "yellow", "part_of_object": "hat"},
        {"id": "N4", "description": "star decoration on hat", "color": "gold", "part_of_object": "hat"},
        {"id": "N5", "description": "left eye pupil", "color": "black", "part_of_object": "left eye"},
        {"id": "N6", "description": "highlight on left eye pupil", "color": "white", "part_of_object": "left eye"},
        {"id": "N7", "description": "lower hat body", "color": "blue", "part_of_object": "hat"}
      ],
      "edges": [
        // Assuming hat body covers head, band covers hat body and divides the hat into upper and lower parts, star covers hat body (or band)
        {"source": "N7", "target": "N1", "relationship": "occludes"},
        {"source": "N3", "target": "N2", "relationship": "occludes"},
        {"source": "N3", "target": "N7", "relationship": "occludes"},
        {"source": "N4", "target": "N2", "relationship": "occludes"}, // Star occludes main hat body
        {"source": "N6", "target": "N5", "relationship": "occludes"}, // Highlight occludes pupil
        {"source": "N7", "target": "N2", "relationship": "interrupted_shape"} // Lower hat body and upper hat body are interrupted shapes
      ]
    }
    ```

* **Non-Occluded Element Analysis from Graph:** `<non_occluded_analysis>...</non_occluded_analysis>`
    Explain which color-region nodes are identified as non-occluded by analyzing the `occludes` relationships in the `<layer_graph>`. Specifically, list the nodes that are not targets in any `occludes(X, Node)` relationship.

* **Caption:** `<caption>...</caption>`
    Provide the final concise caption of *all* identified non-occluded color-region nodes.

**Example:**

---
**Input Image:** (Imagine a cartoon cat. The cat wears a blue hat. A wide red band is on the hat, and a prominent yellow feather is attached, clearly covering the red band. One of the cat's grey ears is visible, but the other is tucked under and occluded by the blue hat. The cat is holding a large green fish, which covers most of its grey body and one grey paw. The cat's face shows a pink nose and black mouth. Its eyes have white sclera, black pupils, and tiny white highlights on the pupils.)

**Output:**

<image_description>
A cartoon cat is shown wearing a blue hat with a red band and a yellow feather. One ear is visible, the other is under the hat. The cat holds a large green fish occluding its body and a paw.
</image_description>

<layer_graph_reasoning>
I segmented the image into distinct flat-color areas.
- Cat: grey body, grey head, grey visible ear, pink nose, black mouth, white sclera (for two eyes), black pupils (for two eyes), small white highlights on pupils (for two eyes), grey visible paw (small part maybe).
- Hat: blue main hat, red band, yellow feather.
- Held item: green fish.
Occlusion analysis was performed as follows:
- The blue hat occludes part of the grey head and one (unseen) grey ear.
- The red band occludes part of the blue hat.
- The yellow feather occludes part of the red band.
- The green fish occludes most of the grey body and one grey paw.
- Each white highlight occludes its respective black pupil.
- Each black pupil occludes its respective white sclera.
</layer_graph_reasoning>

<layer_graph>
```json
{
  "nodes": [
    {"id": "N1", "description": "cat body", "color": "grey", "part_of_object": "cat"},
    {"id": "N2", "description": "cat head", "color": "grey", "part_of_object": "cat"},
    {"id": "N3", "description": "visible cat ear", "color": "grey", "part_of_object":"head"},
    {"id": "N4", "description": "occluded cat ear", "color": "grey", "part_of_object":"head"},
    {"id": "N5", "description": "left eye sclera", "color": "white", "part_of_object": "left eye"},
    {"id": "N6", "description": "right eye sclera", "color": "white", "part_of_object": "right eye"},
    {"id": "N7", "description": "left pupil", "color": "black", "part_of_object": "left eye"},
    {"id": "N8", "description": "right pupil", "color": "black", "part_of_object": "right eye"},
    {"id": "N9", "description": "highlight on left pupil", "color": "white", "part_of_object": "left eye"},
    {"id": "N10", "description": "highlight on right pupil", "color": "white", "part_of_object": "right eye"},
    {"id": "N11", "description": "nose", "color": "pink", "part_of_object": "head"},
    {"id": "N12", "description": "mouth", "color": "black", "part_of_object": "head"},
    {"id": "N13", "description": "main hat body", "color": "blue", "part_of_object": "hat"},
    {"id": "N14", "description": "hat band", "color": "red", "part_of_object": "hat"},
    {"id": "N15", "description": "feather on hat", "color": "yellow", "part_of_object": "hat"},
    {"id": "N16", "description": "held green fish", "color": "green", "part_of_object": "held item"},
    {"id": "N17", "description": "visible cat paw", "color": "grey", "part_of_object": "cat"},
    {"id": "N18", "description": "occluded cat paw", "color": "grey", "part_of_object": "cat"}
  ],
  "edges": [
    {"source": "N13", "target": "N2", "relationship": "occludes"}, // Hat occludes head
    {"source": "N13", "target": "N4", "relationship": "occludes"}, // Hat occludes ear N4
    {"source": "N14", "target": "N13", "relationship": "occludes"},// Band occludes hat body
    {"source": "N15", "target": "N14", "relationship": "occludes"},// Feather occludes band
    {"source": "N16", "target": "N1", "relationship": "occludes"},  // Fish occludes body
    {"source": "N16", "target": "N18", "relationship": "occludes"},// Fish occludes paw N18
    {"source": "N7", "target": "N5", "relationship": "occludes"},   // Pupil occludes sclera
    {"source": "N8", "target": "N6", "relationship": "occludes"},   // Pupil occludes sclera
    {"source": "N9", "target": "N7", "relationship": "occludes"},   // Highlight occludes pupil
    {"source": "N10", "target": "N8", "relationship": "occludes"}  // Highlight occludes pupil
  ]
}
</layer_graph>

<non_occluded_analysis>
Based on the layer graph, the following color-region nodes are not targets in any occludes(X, Node) relationship:
N3 (visible cat ear), N9 (highlight on left pupil), N10 (highlight on right pupil), N11 (nose), N12 (mouth), N15 (feather on hat), N16 (held green fish), and N17 (visible cat paw).
The 'occluded cat ear' (N4) is occluded by N13. 'Left/Right eye sclera' (N5, N6) are occluded by pupils. 'Left/Right pupil' (N7, N8) are occluded by highlights. 'Main hat body' (N13) is occluded by band. 'Hat band' (N14) is occluded by feather. 'Cat body' (N1) is occluded by fish. 'Occluded cat paw' (N18) is occluded by fish.
Therefore, the non-occluded nodes are: N3, N9, N10, N11, N12, N15, N16, N17.
</non_occluded_analysis>
<caption>
The grey visible ear, white highlights on pupils, pink nose, black mouth, yellow feather, green fish, and grey visible paw.
</caption>
\end{lstlisting}

\subsection{Layer Graph Update}
\label{subsection:layer_graph_update}
This system prompt is designed for use after an image editing operation. It instructs the VLM to analyze the current image alongside the layer graph from the previous step, verify and update the graph to accurately reflect the current visual state, and then identify non-occluded elements and generate a caption based on the revised graph.

\begin{lstlisting}
You are an advanced Vision-Language Model tasked with analyzing cartoon-style, flat-color images. You will be given the current image and the layer graph that was constructed for a *previous* version of this image (before an editing operation was applied). Your goal is to:

1.  **Verify, Correct, and Update Layer Graph:** Compare the provided *previous layer graph* with the *current image*.
    *   If the previous layer graph, after any necessary corrections, accurately represents the current image (i.e., the layer removal was successful and the corrected graph reflects the visible elements and their occlusions correctly), you can use this corrected version as a basis.
    *   If the previous layer graph is **no longer accurate** due to changes in the image (e.g., layers were not removed as expected, new artifacts appeared, occlusion relationships have changed), OR if you identified errors in the previous graph itself, you **must update or reconstruct the layer graph** to accurately reflect the *current image*. This might involve adding, removing, or modifying nodes (color regions) and their attributes or relationships (occlusion, interrupted shapes).
    *   Focus on creating a graph that represents the **current visual state** of the image with the highest possible accuracy.

2.  **Layer Graph Construction (if updating/correcting):** If updating or correcting, follow these rules:
    *   **Nodes:** Identify distinct, continuous flat-color regions within the current image. Each such visually separable color region constitutes a node.
    *   **Node Attributes:** For each node (color region), describe:
        *   Its dominant color.
        *   A semantic label if the region clearly corresponds to a recognizable object part.
        *   If no clear semantic label applies, use a geometric description.
        *   `part_of_object(N, O)`: Color-region Node N is part of a larger semantic object O.
    *   **Relationships (Edges):** Define spatial and occlusion relationships:
        *   `occludes(A, B)`: Color-region Node A visually covers Color-region Node B in the current image.
        *   `interrupted_shape(A, B)`: Color-region Node A and B share the same color and are conceptually part of the same larger shape but are visually disconnected in the current image.

3.  **Non-Occluded Element Identification (from updated/corrected graph):** Based *solely* on the `occludes` relationships in the current (potentially updated/corrected) layer graph, identify all color-region nodes that are not occluded by any other node in the *current image*.

4.  **Caption Generation:** Provide a concise caption describing these non-occluded color-region nodes, using their semantic or geometric descriptions.

**Instructions & Definitions:**

*   **Node Granularity:** Each continuous patch of flat color must be a distinct node.
*   **Non-Occluded Element Definition (Graph-based):** A color-region node is non-occluded if no `occludes(X, Node)` relationship exists for it in the *current graph*.
*   **Background Exclusion:** The global white background is not a node.
*   **Internal Details:** White or other colored shapes *inside* objects are distinct color-region nodes.
*   **Concise Captioning:** Group similar non-occluded elements if they belong to the same semantic object. Describe elements using semantic/geometric labels. Avoid 'region' in the caption. Caption < 40 words.

**Input Context (User will provide):**
*   The current image.
*   The layer graph from the *previous* step, formatted as JSON. This graph needs to be checked against the current image, corrected for any existing errors, and updated if necessary.

**Output Format (Strictly Adhere):**

*   **Image Description:** `<image_description>...</image_description>`
    Brief overall description of the *current* image content.

*   **Layer Graph Update Reasoning:** `<layer_graph_reasoning>...</layer_graph_reasoning>`
    Describe your thought process. **Critically, explain if the previous graph was suitable, if and why it needed updates to match the current image, and importantly, if any corrections were made to the previous graph's structure or interpretation based on the current visual evidence, irrespective of the editing operation.** Detail changes made to nodes or edges if any. If no changes were needed beyond validating the previous graph, state that.

*   **Layer Graph (JSON-like Representation):** `<layer_graph>...</layer_graph>`
    Represent the **final, accurate layer graph for the current image**. This will be the validated, corrected, or newly updated one.
    Example structure:
    ```json
    {
      "nodes": [
        {"id": "N1", "description": "...", "color": "...", "part_of_object": "..."}, ...
      ],
      "edges": [
        {"source": "...", "target": "...", "relationship": "..."}, ...
      ]
    }
    ```

*   **Non-Occluded Element Analysis from Graph:** `<non_occluded_analysis>...</non_occluded_analysis>`
    Explain which color-region nodes are identified as non-occluded by analyzing the `occludes` relationships in the `<layer_graph>` you are outputting (the one for the current image).

*   **Caption:** `<caption>...</caption>`
    Provide the final concise caption of *all* identified non-occluded color-region nodes from the *current image*.

**Example Scenario:**
Suppose the previous image showed a "blue square occluding a red circle". The previous graph would be:
Nodes: N1 (blue square), N2 (red circle). Edges: occludes(N1, N2). Non-occluded: N1 (blue square). Caption: "The blue square".
Now, an editing operation attempted to remove the "blue square".
- **If successful:** The current image shows only "red circle".
  Your reasoning: "The blue square was successfully removed. The previous graph is no longer valid. Updated graph contains only the red circle."
  Updated graph: Nodes: N2 (red circle). Edges: []. Non-occluded: N2 (red circle). Caption: "The red circle".
- **If failed:** The current image still shows "blue square occluding red circle".
  Your reasoning: "The editing operation to remove the blue square failed. The previous graph still accurately represents the current image."
  Graph: (Same as previous). Non-occluded: N1 (blue square). Caption: "The blue square".
- **If partially failed / new artifact:** Current image shows "a smaller blue square fragment and the red circle".
  Your reasoning: "The blue square was partially removed. The previous graph needs update to reflect the smaller blue fragment."
  Updated graph: Nodes: N1_frag (blue square fragment), N2 (red circle). Edges: occludes(N1_frag, N2). Non-occluded: N1_frag. Caption: "The blue square fragment".

Proceed with the analysis of the provided current image, considering the provided previous layer graph for verification, correction, and updates.
"""
\end{lstlisting}

\subsection{Instance Mask, Box, and Prompt Generation}
\label{subsection:box_detection}
Given a specification of target elements or layers, this system prompt directs the VLM to output segmentation masks, 2D bounding boxes, and descriptive text labels (serving as instance prompts) for those specific visual components. This prompt is adapted from an official prompt structure provided by Google Gemini.
\begin{lstlisting}
Give the segmentation masks for the "{layers}". Output a JSON list of segmentation masks where each entry contains the 2D bounding box in the key "box_2d", the segmentation mask in key "mask", and the text label in the key "label". Use descriptive labels.
\end{lstlisting}

\subsection{Isolated Layer Annotation}
\label{subsection:layer_annotation}
This system prompt is used during dataset construction. It provides the VLM with a split image showing the original context (Part A) and isolated visual elements (Part B), and instructs it to analyze both parts to generate a descriptive annotation for the isolated elements in Part B, leveraging the contextual information from Part A.
\begin{lstlisting}
You are provided an image (width 1024, height 512) annotated with letters "A" and "B" in the top left corner, with a black vertical line at x=512, separating the image into two parts:
- **Part A:** Occupies [0:512, 0:512], with letter "A" in the top left corner. Shows the original icon context.
- **Part B:** Occupies [512:1024, 0:512], with letter "B" in the top left corner. Shows *only* specific elements isolated from the original icon.

**Your Task:** Analyze both parts of the image and generate a response containing a caption, your thinking process, and a detailed description of the elements in Part B, following the format below.
1.  **Analyze Context (Part A):** Understand the original context of the elements shown in Part B by examining Part A.
2.  **Identify Elements (Part B):** Carefully identify each distinct element present in Part B.
3.  **Describe Elements (using context from A):**
    *   For *each* element identified in Part B, determine its role or identity using Part A.
    *   If an element has a clear semantic role in Part A, describe the element using its **meaning** and key visual properties (e.g., "the blue left eye," "the central gear component," "the red curved mouth"). Be definitive.
    *   If the element's semantic meaning is vague or not clear in Part A, describe the element primarily by its **geometric and visual properties** (e.g., "the black triangle on the left," "the white pattern inside a green circle").
    *   The description should be accurate enough that someone could locate the described element within Part A, even without seeing Part B (e.g., describing "the grey cloud in front of the sun" from Part B makes it findable in Part A, better than just saying "the cloud").

# Output Format:

*   **Caption:** Within `<caption>...</caption>` tags, write a detailedcaption (30-50 words) for both Part A and Part B.
    *   Describe the object, elements, labels, etc. shown in Part A (appearance, main parts, colors). If Part A's semantic meaning is vague, do not guess the potential meaning.
    *   State what Part B shows.
    *   Do not mention the transparent background in Part A and B.

*   **Thinking Process:** Within `<thinking>...</thinking>` tags:
    *   Briefly explain your interpretation of Part A.
    *   List the distinct elements you identified in Part B.
    *   For each element, explain how you used (or why you couldn't use) Part A to arrive at its description (semantic vs. geometric choice).
    *   For elements in Part B that are parts of a larger object or have similar counterparts in Part A, include spatial relationships to aid localization (e.g., "two triangles at the top-left and top-right of the diamond", "three parallel line segments of the letter 'E'", "the leftmost circle of three identical circles").

*   **Description:** Within `<description>...</description>` tags:
    *   Provide the final, concise description summarizing *all* identified elements in Part B, based on your analysis above.
    *   List elements in Part B as comma-separated phrases. Use noun phrases only - avoid verbs like "is", "are", "has", etc.
    *   IMPORTANT: Top-left letters "A" and "B" do not belong to the image content. Do not include letters "A" or "B" in the description text.
    *   Some elements can be grouped together for better readability without losing any information. For example, "Upper teal bar, lower teal bar, left teal bar, center teal bar, right teal bar" can be grouped together as "teal bars"; "Outline of the laptop screen, outline of the laptop base, outline of the downward-pointing arrow" can be grouped together as "outline of the laptop".

**Example 1 (Sufficient Context in A):**
###
<caption>Part A (Left) shows a cartoon face icon. Part B (Right) contains a blue circle and a red curved line.</caption>
<thinking>
Part A shows a recognizable, though incomplete, cartoon face.
Identified elements in B: 1. A small filled blue circle. 2. A simple red curved line.
Context mapping: The blue circle corresponds to the left eye in Part A. The red curve corresponds to the mouth. Both have clear semantic roles.
Avoid including part "A" or "B" in the description text.
</thinking>
<description>The left blue eye and the red curved mouth.</description>
###

**Example 2 (Insufficient Context / Abstract):**
###
<caption>Part A (Left) shows an abstract graphic with overlapping black and green shapes. Part B (Right) has a black rectangular shape.</caption>
<thinking>
Part A is abstract.
Identified elements in B: A black rectangle.
Context mapping: The rectangle has no clear semantic role in the abstract graphic in Part A. Describing geometrically.
Avoid including part "A" or "B" in the description text.
</thinking>
<description>The black rectangle over the green circle.</description>
###

**Example 3 (Very Incomplete Context / Identical Parts):**
###
<caption>Part A shows a green outline of a rabbit figure. Part B shows the same green outline.</caption>
<thinking>
Part A and B are identical, showing only the green outline of a rabbit.
Avoid including part "A" or "B" in the description text.
</thinking>
<description>The green outline of a rabbit.</description>
###

Return your response in the above format without preambles like "OK, I understand..."
\end{lstlisting}

\subsection{Ablation Study: Direct Top Layer Detection}
\label{subsection:ablation_prompt}
This system prompt is specifically designed for an ablation study. It instructs the VLM to identify visual elements that appear entirely non-occluded based only on visual overlaps and perceived stacking in the rendered image, without relying on a constructed layer graph, to serve as a baseline for comparison.
\begin{lstlisting}
Analyze the provided flat-color, cartoon-style image. Based purely on the visual presentation, infer the visual stacking and occlusion of elements. Your goal is to identify **only** the elements that are visually positioned at the very top, meaning they are **not covered by any other element** within the image. Provide a concise caption describing these non-occluded elements.

# Core Task:
Identify *all* visual elements that are completely unobstructed by *any other element* in the image, based on visual analysis of overlaps and occlusion. These are the non-occluded visual elements.

# Definition of Non-Occluded Element:
An element is considered non-occluded if and only if no other element in the image is visually positioned above *any part* of it. This must be determined by analyzing how elements visually overlap and obscure each other in the rendered image.

# Instructions for Description and Caption:
1.  **Semantic Description:** If a non-occluded element has a clear real-world meaning (e.g., "eye", "hat", "wheel", "liquid"), describe it using its **meaning** and key visual attributes (e.g., "the green left eye," "the tall blue hat", "the black outline of the computer screen", "the light blue texture over the bottle body").
2.  **Geometric Description:** If a non-occluded element's meaning is abstract or unclear, describe it using its primary **geometric shape, color, and relative position** (e.g., "the large red circle in the center", "the thin black outline curving upwards on the right").
3.  The white background filling the entire canvas is not considered a specific element. Ignore it.
4.  White shapes *inside* other shapes (e.g., liquid, highlights, inner details) are separate elements and *should* be considered if they are non-occluded and visible.
5.  **Object Count Limit:** If there are too many non-occluded elements (e.g., >= 5), consider grouping the elements to describe them collectively (e.g., "the black left eye, the black right eye" -> "the black eyes"), or return only a subset of them based on similarity (i.e., position, color, shape, etc.).

# Output Format:
*   **Description:** In `<description>...</description>` tags, provide a concise description of the overall image content or key visible components. This sets the context.
*   **Think Process:** Enclose your reasoning within `<think>...</think>` tags. **Describe your visual analysis process.** Identify the distinct visual elements and **analyze how they overlap and obscure each other** to infer the visual stacking order. Based on this visual stacking determined from occlusion, identify the element(s) that are not covered by *any* other element. Explain why elements that are partially or fully covered by others are excluded. For example, if element A is visually covering element B, B is not non-occluded. If B is covering C, C is not non-occluded. Only elements with *no* part visually obscured by *any other element* are non-occluded.
*   **Caption:** Enclose the final description of *all* identified non-occluded element(s) within `<caption>...</caption>` tags. This should be a list or combined description of the non-occluded elements found. The caption should be LESS than 40 words. To reduce word count, you may need to group similar elements or describe them collectively.

# Example 0 (Semantic Element):
###
<description>
The image shows a simple cartoon face. The face has two black circular eyes positioned symmetrically, a curved red line forming a smile, and a yellow star-shaped decoration placed on top of the head. Two hands with darker color curves cover portions of the face.
</description>
<think>
I am analyzing the visual image for elements and how they overlap. I see distinct shapes: the face, eyes, mouth, star, and hands. I observe that the hands cover parts of the face. However, the eyes, mouth, and star are fully visible and no part of them is covered by the hands or any other element in the image. The hands themselves are not single-color continuous shapes, they have darker color curves inside them. So hands are not non-occluded.
</think>
<caption>The black left eye, the black right eye, the red mouth, and the yellow star on the hair</caption>
###

# Example 1 (Semantic Element - Partial Covering):
###
<description>
The image contains a pink peach with two green leaves and a dark-red curve on the peach body. The right leaf is partially covered by the peach body.
</description>
<think>
I am analyzing the visual image for overlaps. I see the peach, the dark-red curve on the peach, and two leaves. The dark-red curve is fully visible on the peach. The left leaf is fully visible. The right leaf has a portion of its shape obscured by the peach body. Therefore, the dark-red curve and the left leaf are the elements not covered by any other element.
</think>
<caption>The dark-red curve on the peach body and the left leaf</caption>
###

# Example 2 (Nested Covering - Abstract Elements):
###
<description>
The image contains a black circle visually positioned over a green rectangle, which is visually positioned over a yellow triangle.
</description>
<think>
I am analyzing the visual image for overlaps. I see a black circle, a green rectangle, and a yellow triangle. The black circle visually covers the green rectangle. The green rectangle visually covers the yellow triangle. This establishes a visual stacking order: Circle > Rectangle > Triangle. The black circle is not visually covered by any other element. The green rectangle is covered by the black circle, and the yellow triangle is covered by the green rectangle. Therefore, only the black circle is non-occluded.
</think>
<caption>The black circle in the center</caption>
###

# Example 3 (Multiple Unrelated Non-Occluded Elements):
###
<description>
The image shows a small blue square in the top-left corner and a large yellow triangle dominating the center. These two elements are visually separate and do not overlap.
</description>
<think>
I am analyzing the visual image for overlaps. I see a small blue square and a large yellow triangle. These two elements do not visually overlap at all. The yellow triangle is fully visible and not covered by any other element. The small blue square is also fully visible and not covered by any other element. Both meet the criteria for being non-occluded elements as nothing obscures them.
</think>
<caption>The large yellow triangle and the small blue square in the top-left</caption>
###

Proceed with the analysis of the provided image.
\end{lstlisting}

%% file: ref.bib
@String{Computer = "{IEEE} Computer" }

@inproceedings{Li:2020:DVG,
    title={Differentiable Vector Graphics Rasterization for Editing and Learning},
    author={Li, Tzu-Mao and Luk\'{a}\v{c}, Michal and Gharbi Micha\"{e}l and Jonathan Ragan-Kelley},
    journal={ACM Transactions on Graphics},
    volume={39},
    number={6},
    pages={1--15},
    year={2020}
}

@inproceedings{ma2022towards,
    title={Towards Layer-wise Image Vectorization},
    author={Ma, Xu and Zhou, Yuqian and Xu, Xingqian and Sun, Bin and Filev, Valerii and Orlov, Nikita and Fu, Yun and Shi, Humphrey},
    booktitle={IEEE Conference on Computer Vision and Pattern Recognition},
    pages={16314--16323},
    year={2022}
}

@inproceedings{reddy2021im2vec,
    title={{Im2Vec}: {Synthesizing} Vector Graphics without Vector Supervision},
    author={Reddy, Pradyumna and Gharbi, Michael and Lukac, Michal and Mitra, Niloy J.},
    booktitle={IEEE Conference on Computer Vision and Pattern Recognition},
    pages={7342--7351},
    year={2021}
}

@inproceedings{huang2025photodoodle,
  title={{PhotoDoodle}: {Learning} Artistic Image Editing from Few-Shot Pairwise Data},
  author={Huang, Shijie and Song, Yiren and Zhang, Yuxuan and Guo, Hailong and Wang, Xueyin and Shou, Mike Zheng and Liu, Jiaming},
  booktitle={arXiv preprint arXiv:2502.14397},
  year={2025}
}

@inproceedings{ge2024seed,
  title={{SEED-Data-Edit Technical Report}: {A} Hybrid Dataset for Instructional Image Editing},
  author={Ge, Yuying and Zhao, Sijie and Li, Chen and Ge, Yixiao and Shan, Ying},
  booktitle={arXiv preprint arXiv:2405.04007},
  year={2024}
}

@inproceedings{hirschorn2024optimize,
  title={{Optimize \& Reduce}: A Top-Down Approach for Image Vectorization},
  author={Hirschorn, Or and Jevnisek, Amir and Avidan, Shai},
  booktitle={AAAI Conference on Artificial Intelligence},
  pages={2148--2156},
  year={2024}
}

@inproceedings{zhou2024segmentation,
  title={Segmentation-Guided Layer-Wise Image Vectorization with Gradient Fills},
  author={Zhou, Hengyu and Zhang, Hui and Wang, Bin},
  booktitle={European Conference on Computer Vision},
  pages={165--180},
  year={2024},
}

@inproceedings{wang2024layered,
  title={Layered Image Vectorization via Semantic Simplification},
  author={Wang, Zhenyu and Huang, Jianxi and Sun, Zhida and Gong, Yuanhao and Cohen-Or, Daniel and Lu, Min},
  booktitle={arXiv preprint arXiv:2406.05404},
  year={2024}
}

@inproceedings{song2025layertracer,
  title={{LayerTracer}: {Cognitive-Aligned} Layered SVG Synthesis via Diffusion Transformer},
  author={Song, Yiren and Chen, Danze and Shou, Mike Zheng},
  booktitle={arXiv preprint arXiv:2502.01105},
  year={2025}
}

@inproceedings{chen2023editable,
  title={Editable Image Geometric Abstraction via Neural Primitive Assembly},
  author={Chen, Ye and Ni, Bingbing and Chen, Xuanhong and Hu, Zhangli},
  booktitle={IEEE International Conference on Computer Vision},
  pages={23514--23523},
  year={2023}
}

@inproceedings{thamizharasan2024vecfusion,
  title={{VecFusion}: {Vector} Font Generation with Diffusion},
  author={Thamizharasan, Vikas and Liu, Difan and Agarwal, Shantanu and Fisher, Matthew and Gharbi, Micha{\"e}l and Wang, Oliver and Jacobson, Alec and Kalogerakis, Evangelos},
  booktitle={IEEE Conference on Computer Vision and Pattern Recognition},
  pages={7943--7952},
  year={2024}
}

@inproceedings{shen2021clipgen,
  title={{ClipGen}: {A} Deep Generative Model for Clipart Vectorization and Synthesis},
  author={Shen, I-Chao and Chen, BingYu},
  journal={IEEE Transactions on Visualization and Computer Graphics},
  volume={28},
  number={12},
  pages={4211--4224},
  year={2021},
}

@inproceedings{poole2022dreamfusion,
  title={{Dreamfusion}: {Text-to-3D} using 2D Diffusion},
  author={Poole, Ben and Jain, Ajay and Barron, Jonathan T. and Mildenhall, Ben},
  booktitle={arXiv preprint arXiv:2209.14988},
  year={2022}
}

@inproceedings{hu2021lora,
  title={{LoRA}: {Low-Rank} Adaptation of Large Language Models},
  author={Hu, Edward J. and Shen, Yelong and Wallis, Phillip and Allen-Zhu, Zeyuan and Li, Yuanzhi and Wang, Shean and Wang, Lu and Chen, Weizhu},
  booktitle={arXiv preprint arXiv:2106.09685},
  year={2021}
}

@inproceedings{peebles2023scalable,
  title={Scalable Diffusion Models with Transformers},
  author={Peebles, William and Xie, Saining},
  booktitle={IEEE International Conference on Computer Vision},
  pages={4195--4205},
  year={2023}
}

@inproceedings{raffel2020exploring,
  title={Exploring the Limits of Transfer Learning with a Unified Text-to-Text Transformer},
  author={Raffel, Colin and Shazeer, Noam and Roberts, Adam and Lee, Katherine and Narang, Sharan and Matena, Michael and Zhou, Yanqi and Li, Wei and Liu, Peter J.},
  journal={Journal of Machine Learning Research},
  volume={21},
  number={140},
  pages={1--67},
  year={2020}
}

@inproceedings{esser2024scaling,
  title={Scaling Rectified Flow Transformers for High-Resolution Image Synthesis},
  author={Esser, Patrick and Kulal, Sumith and Blattmann, Andreas and Entezari, Rahim and M{\"u}ller, Jonas and Saini, Harry and Levi, Yam and Lorenz, Dominik and Sauer, Axel and Boesel, Frederic and Podell, Dustin and Dockhorn, Tim and English, Zion and Lacey, Kyle and Goodwin, Alex and Marek, Yannik and Rombach, Robin},
  booktitle={International Conference on Machine Learning},
  year={2024}
}

@inproceedings{law2025image,
  title={{Image Vectorization with Depth}: Convexified Shape Layers with Depth Ordering},
  author={Law, Ho and Kang, SungHa},
  journal={SIAM Journal on Imaging Sciences},
  volume={18},
  number={2},
  pages={963--1001},
  year={2025},
}

@inproceedings{zhang2018unreasonable,
  title={The Unreasonable Effectiveness of Deep Features as a Perceptual Metric},
  author={Zhang, Richard and Isola, Phillip and Efros, Alexei A. and Shechtman, Eli and Wang, Oliver},
  booktitle={IEEE Conference on Computer Vision and Pattern Recognition},
  pages={586--595},
  year={2018}
}

@inproceedings{simonyan2014very,
  title={Very Deep Convolutional Networks for Large-scale Image Recognition},
  author={Simonyan, Karen and Zisserman, Andrew},
  booktitle={arXiv preprint arXiv:1409.1556},
  year={2014}
}

@inproceedings{radford2021learning,
  title={Learning Transferable Visual Models From Natural Language Supervision},
  author={Radford, Alec and Kim, Jong-Wook and Hallacy, Chris and Ramesh, Aditya and Goh, Gabriel and Agarwal, Sandhini and Sastry, Girish and Askell, Amanda and Mishkin, Pamela and Clark, Jack and Krueger, Gretchen and Sutskever, Ilya},
  booktitle={International Conference on Machine Learning},
  pages={8748--8763},
  year={2021},
}

@inproceedings{tian2022survey,
  title={{A Survey of Smooth Vector Graphics}: {Recent} Advances in Representation, Creation, Rasterization, and Image Vectorization},
  author={Tian, Xingze and G{\"u}nther, Tobias},
  journal={IEEE Transactions on Visualization and Computer Graphics},
  volume={30},
  number={3},
  pages={1652--1671},
  year={2022},
}

@inproceedings{battiato2004svg,
  title={SVG Rendering of Real Images Using Data Dependent Triangulation},
  author={Battiato, Sebastiano and Gallo, Giovanni and Messina, Giuseppe},
  booktitle={Spring Conference on Computer Graphics},
  pages={185--192},
  year={2004}
}

@inproceedings{demaret2006image,
  title={Image Compression by Linear Splines over Adaptive Triangulations},
  author={Demaret, Laurent and Dyn, Nira and Iske, Armin},
  journal={Signal Processing},
  volume={86},
  number={7},
  pages={1604--1616},
  year={2006},
}

@inproceedings{swaminarayan2006rapid,
  title={Rapid Automated Polygonal Image Decomposition},
  author={Swaminarayan, Sriram and Prasad, Lakshman},
  booktitle={IEEE Applied Imagery and Pattern Recognition Workshop},
  pages={28--28},
  year={2006},
}

@inproceedings{lecot2006ardeco,
  title={{ARDECO}: {Automatic} Region DEtection and Conversion},
  author={Lecot, Gregory and Levy, Bruno},
  booktitle={Eurographics Symposium on Rendering},
  pages={349--360},
  year={2006}
}

@inproceedings{liao2012subdivision,
  title={A Subdivision-Based Representation for Vector Image Editing},
  author={Liao, Zicheng and Hoppe, Hugues and Forsyth, David and Yu, Yizhou},
  journal={IEEE Transactions on Visualization and Computer Graphics},
  volume={18},
  number={11},
  pages={1858--1867},
  year={2012},
}

@inproceedings{xia2009patch,
  title={Patch-Based Image Vectorization with Automatic Curvilinear Feature Alignment},
  author={Xia, Tian and Liao, Binbin and Yu, Yizhou},
  journal={ACM Transactions on Graphics},
  volume={28},
  number={5},
  pages={1--10},
  year={2009},
}

@inproceedings{yang2015effective,
  title={Effective Clipart Image Vectorization Through Direct Optimization of Bezigons},
  author={Yang, Ming and Chao, Hongyang and Zhang, Chi and Guo, Jun and Yuan, Lu and Sun, Jian},
  journal={IEEE Transactions on Visualization and Computer Graphics},
  volume={22},
  number={2},
  pages={1063--1075},
  year={2015},
}

@inproceedings{sun2007image,
  title={Image Vectorization Using Optimized Gradient Meshes},
  author={Sun, Jian and Liang, Lin and Wen, Fang and Shum, Heung-Yeung},
  journal={ACM Transactions on Graphics},
  volume={26},
  number={3},
  pages={11--es},
  year={2007},
}

@inproceedings{orzan2008diffusion,
  title={{Diffusion Curves}: {A} Vector Representation for Smooth-Shaded Images},
  author={Orzan, Alexandrina and Bousseau, Adrien and Winnem{\"o}ller, Holger and Barla, Pascal and Thollot, Jo{\"e}lle and Salesin, David},
  journal={ACM Transactions on Graphics},
  volume={27},
  number={3},
  pages={1--8},
  year={2008},
}

@inproceedings{xie2014hierarchical,
  title={Hierarchical Diffusion Curves for Accurate Automatic Image Vectorization},
  author={Xie, Guofu and Sun, Xin and Tong, Xin and Nowrouzezahrai, Derek},
  journal={ACM Transactions on Graphics},
  volume={33},
  number={6},
  pages={1--11},
  year={2014},
}

@inproceedings{zhao2017inverse,
  title={Inverse Diffusion Curves using Shape Optimization},
  author={Zhao, Shuang and Durand, Fr{\'e}do and Zheng, Changxi},
  journal={IEEE Transactions on Visualization and Computer Graphics},
  volume={24},
  number={7},
  pages={2153--2166},
  year={2017},
}

@inproceedings{brooks2023instructpix2pix,
  title={{InstructPix2Pix}: {Learning} to Follow Image Editing Instructions},
  author={Brooks, Tim and Holynski, Aleksander and Efros, Alexei A.},
  booktitle={IEEE Conference on Computer Vision and Pattern Recognition},
  pages={18392--18402},
  year={2023}
}

@inproceedings{wei2024omniedit,
  title={{OmniEdit}: {Building} Image Editing Generalist Models Through Specialist Supervision},
  author={Wei, Cong and Xiong, Zheyang and Ren, Weiming and Du, Xeron and Zhang, Ge and Chen, Wenhu},
  booktitle={International Conference on Learning Representations},
  year={2024}
}

@inproceedings{xu2024insightedit,
  title={{InsightEdit}: {Towards} Better Instruction Following for Image Editing},
  author={Xu, Yingjing and Kong, Jie and Wang, Jiazhi and Pan, Xiao and Lin, Bo and Liu, Qiang},
  booktitle={arXiv preprint arXiv:2411.17323},
  year={2024}
}

@inproceedings{yu2024anyedit,
  title={{AnyEdit}: {Mastering} Unified High-Quality Image Editing for Any Idea},
  author={Yu, Qifan and Chow, Wei and Yue, Zhongqi and Pan, Kaihang and Wu, Yang and Wan, Xiaoyang and Li, Juncheng and Tang, Siliang and Zhang, Hanwang and Zhuang, Yueting},
  booktitle={arXiv preprint arXiv:2411.15738},
  year={2024}
}

@inproceedings{zhang2023magicbrush,
  title={{MagicBrush}: {A} Manually Annotated Dataset for Instruction-Guided Image Editing},
  author={Zhang, Kai and Mo, Lingbo and Chen, Wenhu and Sun, Huan and Su, Yu},
  journal={Advances in Neural Information Processing Systems},
  pages={31428--31449},
  year={2023}
}

@inproceedings{zhao2024ultraedit,
  title={{UltraEdit}: {Instruction-based} Fine-Grained Image Editing at Scale},
  author={Zhao, Haozhe and Ma, Xiaojian Shawn and Chen, Liang and Si, Shuzheng and Wu, Rujie and An, Kaikai and Yu, Peiyu and Zhang, Minjia and Li, Qing and Chang, Baobao},
  journal={Advances in Neural Information Processing Systems},
  volume={37},
  pages={3058--3093},
  year={2024}
}

@inproceedings{dong2023prompt,
  title={Prompt Tuning Inversion for Text-Driven Image Editing Using Diffusion Models},
  author={Dong, Wenkai and Xue, Song and Duan, Xiaoyue and Han, Shumin},
  booktitle={IEEE International Conference on Computer Vision},
  pages={7430--7440},
  year={2023}
}

@inproceedings{hertz2022prompt,
  title={Prompt-to-Prompt Image Editing with Cross Attention Control},
  author={Hertz, Amir and Mokady, Ron and Tenenbaum, Jay and Aberman, Kfir and Pritch, Yael and Cohen-Or, Daniel},
  booktitle={arXiv preprint arXiv:2208.01626},
  year={2022}
}

@inproceedings{meng2021sdedit,
  title={{SDEdit}: {Guided} Image Synthesis and Editing with Stochastic Differential Equations},
  author={Meng, Chenlin and He, Yutong and Song, Yang and Song, Jiaming and Wu, Jiajun and Zhu, Jun-Yan and Ermon, Stefano},
  booktitle={arXiv preprint arXiv:2108.01073},
  year={2021}
}

@inproceedings{tumanyan2023plug,
  title={Plug-and-Play Diffusion Features for Text-Driven Image-to-Image Translation},
  author={Tumanyan, Narek and Geyer, Michal and Bagon, Shai and Dekel, Tali},
  booktitle={IEEE Conference on Computer Vision and Pattern Recognition},
  pages={1921--1930},
  year={2023}
}

@inproceedings{avrahami2024stable,
  title={{Stable Flow}: {Vital} Layers for Training-Free Image Editing},
  author={Avrahami, Omri and Patashnik, Or and Fried, Ohad and Nemchinov, Egor and Aberman, Kfir and Lischinski, Dani and Cohen-Or, Daniel},
  booktitle={arXiv preprint arXiv:2411.14430},
  year={2024}
}

@inproceedings{cao2023masactrl,
  title={{MasaCtrl}: {Tuning-Free} Mutual Self-Attention Control for Consistent Image Synthesis and Editing},
  author={Cao, Mingdeng and Wang, Xintao and Qi, Zhongang and Shan, Ying and Qie, Xiaohu and Zheng, Yinqiang},
  booktitle={IEEE International Conference on Computer Vision},
  pages={22560--22570},
  year={2023}
}

@inproceedings{li2023stylediffusion,
  title={{StyleDiffusion}: {Prompt-Embedding} Inversion for Text-Based Editing},
  author={Li, Senmao and Van De Weijer, Joost and Hu, Taihang and Khan, Fahad Shahbaz and Hou, Qibin and Wang, Yaxing and Yang, Jian},
  booktitle={arXiv preprint arXiv:2303.15649},
  year={2023}
}

@inproceedings{tewel2024add,
  title={{Add-it}: {Training-Free} Object Insertion in Images With Pretrained Diffusion Models},
  author={Tewel, Yoad and Gal, Rinon and Samuel, Dvir and Atzmon, Yuval and Wolf, Lior and Chechik, Gal},
  booktitle={arXiv preprint arXiv:2411.07232},
  year={2024}
}

@inproceedings{xu2024headrouter,
  title={{HeadRouter}: {A} Training-free Image Editing Framework for MM-DiTs by Adaptively Routing Attention Heads},
  author={Xu, Yu and Tang, Fan and Cao, Juan and Zhang, Yuxin and Kong, Xiaoyu and Li, Jintao and Deussen, Oliver and Lee, Tong-Yee},
  booktitle={arXiv preprint arXiv:2411.15034},
  year={2024}
}

@inproceedings{ju2023direct,
  title={{Direct Inversion}: {Boosting} Diffusion-based Editing with 3 Lines of Code},
  author={Ju, Xuan and Zeng, Ailing and Bian, Yuxuan and Liu, Shaoteng and Xu, Qiang},
  booktitle={arXiv preprint arXiv:2310.01506},
  year={2023}
}

@inproceedings{lin2024schedule,
  title={{Schedule Your Edit}: {A} Simple yet Effective Diffusion Noise Schedule for Image Editing},
  author={Lin, Haonan and Chen, Yan and Wang, Jiahao and An, Wenbin and Wang, Mengmeng and Tian, Feng and Liu, Yong and Dai, Guang and Wang, Jingdong and Wang, Qianying},
  booktitle={Advances in Neural Information Processing Systems},
  pages={115712--115756},
  year={2024}
}

@inproceedings{miyake2025negative,
  title={{Negative-prompt Inversion}: {Fast} Image Inversion for Editing with Text-guided Diffusion Models},
  author={Miyake, Daiki and Iohara, Akihiro and Saito, Yu and Tanaka, Toshiyuki},
  booktitle={IEEE Winter Conference on Applications of Computer Vision},
  pages={2063--2072},
  year={2025},
}

@inproceedings{mokady2023null,
  title={Null-text Inversion for Editing Real Images using Guided Diffusion Models},
  author={Mokady, Ron and Hertz, Amir and Aberman, Kfir and Pritch, Yael and Cohen-Or, Daniel},
  booktitle={IEEE Conference on Computer Vision and Pattern Recognition},
  pages={6038--6047},
  year={2023}
}

@inproceedings{wang2024taming,
  title={Taming Rectified Flow for Inversion and Editing},
  author={Wang, Jiangshan and Pu, Junfu and Qi, Zhongang and Guo, Jiayi and Ma, Yue and Huang, Nisha and Chen, Yuxin and Li, Xiu and Shan, Ying},
  journal={arXiv preprint arXiv:2411.04746},
  year={2024}
}

@inproceedings{dominici2020polyfit,
  title={{PolyFit}: {Perception-aligned} Vectorization of Raster Clip-art via Intermediate Polygonal Fitting},
  author={Dominici, Edoardo Alberto and Schertler, Nico and Griffin, Jonathan and Hoshyari, Shayan and Sigal, Leonid and Sheffer, Alla},
  journal={ACM Transactions on Graphics},
  volume={39},
  number={4},
  pages={77--1},
  year={2020},
}

@inproceedings{du2023image,
  title={Image Vectorization and Editing via Linear Gradient Layer Decomposition},
  author={Du, Zhengjun and Kang, Liangfu and Tan, Jianchao and Gingold, Yotam and Xu, Kun},
  journal={ACM Transactions on Graphics},
  volume={42},
  number={4},
  pages={1--13},
  year={2023},
}

@inproceedings{favreau2017photo2clipart,
  title={{Photo2ClipArt}: {Image} Abstraction and Vectorization Using Layered Linear Gradients},
  author={Favreau, Jean-Dominique and Lafarge, Florent and Bousseau, Adrien},
  journal={ACM Transactions on Graphics},
  volume={36},
  number={6},
  pages={1--11},
  year={2017},
}

@inproceedings{hoshyari2018perception,
  title={Perception-Driven Semi-Structured Boundary Vectorization},
  author={Hoshyari, Shayan and Dominici, Edoardo Alberto and Sheffer, Alla and Carr, Nathan and Wang, Zhaowen and Ceylan, Duygu and Shen, I-Chao},
  journal={ACM Transactions on Graphics},
  volume={37},
  number={4},
  pages={1--14},
  year={2018},
}

@inproceedings{yan2024deep,
  title={Deep Sketch Vectorization via Implicit Surface Extraction},
  author={Yan, Chuan and Li, Yong and Aneja, Deepali and Fisher, Matthew and Simo-Serra, Edgar and Gingold, Yotam},
  journal={ACM Transactions on Graphics},
  volume={43},
  number={4},
  pages={1--13},
  year={2024},
}

@inproceedings{Zhang2009VectorizingCA,
  title={Vectorizing Cartoon Animations},
  author={Song-Hai Zhang and Tao Chen and Yi-Fei Zhang and Shimin Hu and Ralph Robert M.},
  journal={IEEE Transactions on Visualization and Computer Graphics},
  year={2009},
  volume={15},
  number={4},
  pages={618-629},
}

@inproceedings{su2021marvel,
  title={{MARVEL}: {Raster} Manga Vectorization via Primitive-wise Deep Reinforcement Learning},
  author={Su, Hao and Niu, Jianwei and Liu, Xuefeng and Cui, Jiahe and Wan, Ji},
  journal={arXiv preprint arXiv:2110.04830},
  year={2021}
}

@inproceedings{kopf2011depixelizing,
  title={Depixelizing Pixel Art},
  author={Kopf, Johannes and Lischinski, Dani},
  booktitle={SIGGRAPH},
  pages={1--8},
  year={2011}
}

@inproceedings{yang2024posterllava,
  title={{PosterLLaVa}: {Constructing} a Unified Multi-modal Layout Generator with {LLM}},
  author={Yang, Tao and Luo, Yingmin and Qi, Zhongang and Wu, Yang and Shan, Ying and Chen, Chang Wen},
  booktitle={arXiv preprint arXiv:2406.02884},
  year={2024}
}

@inproceedings{cheng2024graphic,
  title={Graphic Design with Large Multimodal Model},
  author={Cheng, Yutao and Zhang, Zhao and Yang, Maoke and Nie, Hui and Li, Chunyuan and Wu, Xinglong and Shao, Jie},
  booktitle={arXiv preprint arXiv:2404.14368},
  year={2024}
}

@inproceedings{sun2024layoutvlm,
  title={{LayoutVLM}: {Differentiable} Optimization of 3D Layout via Vision-Language Models},
  author={Sun, Fan-Yun and Liu, Weiyu and Gu, Siyi and Lim, Dylan and Bhat, Goutam and Tombari, Federico and Li, Manling and Haber, Nick and Wu, Jiajun},
  booktitle={arXiv preprint arXiv:2412.02193},
  year={2024}
}

@inproceedings{ganeshan2024parsel,
  title={{ParSEL}: {Parameterized} Shape Editing with Language},
  author={Ganeshan, Aditya and Huang, Ryan Y. and Xu, Xianghao and Jones, R. Kenny and Ritchie, Daniel},
  booktitle={arXiv preprint arXiv:2405.20319},
  year={2024}
}

@inproceedings{huang2024blenderalchemy,
  title={{BlenderAlchemy}: {Editing} {3D} Graphics with Vision-Language Models},
  author={Huang, Ian and Yang, Guandao and Guibas, Leonidas},
  booktitle={arXiv preprint arXiv:2404.17672},
  year={2024}
}

@inproceedings{liu2024logomotion,
  title={{LogoMotion}: {Visually} Grounded Code Generation for Content-Aware Animation},
  author={Liu, Vivian and Kazi, Rubaiat H. and Wei, Li-Yi and Fisher, Matthew and Langlois, Timothy and Walker, Seth and Chilton, Lydia},
  booktitle={arXiv preprint arXiv:2405.07065},
  year={2024}
}

@inproceedings{ma2025mover,
  title={{MoVer}: {Motion} Verification for Motion Graphics Animations},
  author={Ma, Jiaju and Agrawala, Maneesh},
  booktitle={arXiv preprint arXiv:2502.13372},
  year={2025}
}

@inproceedings{kulits2024re,
  title={Re-Thinking Inverse Graphics With Large Language Models},
  author={Kulits, Peter and Feng, Haiwen and Liu, Weiyang and Abrevaya, Victoria and Black, Michael J.},
  booktitle={arXiv preprint arXiv:2404.15228},
  year={2024}
}

@inproceedings{wu2024gpt,
  title={{GPT-4V} (ision) is a human-aligned evaluator for text-to-{3D} generation},
  author={Wu, Tong and Yang, Guandao and Li, Zhibing and Zhang, Kai and Liu, Ziwei and Guibas, Leonidas and Lin, Dahua and Wetzstein, Gordon},
  booktitle={IEEE Conference on Computer Vision and Pattern Recognition},
  pages={22227--22238},
  year={2024}
}

@inproceedings{yang2023set,
  title={Set-of-Mark Prompting Unleashes Extraordinary Visual Grounding in GPT-4V},
  author={Yang, Jianwei and Zhang, Hao and Li, Feng and Zou, Xueyan and Li, Chunyuan and Gao, Jianfeng},
  booktitle={arXiv preprint arXiv:2310.11441},
  year={2023}
}

@inproceedings{cai2024vip,
  title={ViP-LLaVA: Making Large Multimodal Models Understand Arbitrary Visual Prompts},
  author={Cai, Mu and Liu, Haotian and Mustikovela, Siva K. and Meyer, Gregory P. and Chai, Yuning and Park, Dennis and Lee, Yong Jae},
  booktitle={IEEE/CVF Conference on Computer Vision and Pattern Recognition},
  pages={12914--12923},
  year={2024}
}

@inproceedings{lei2024scaffolding,
  title={Scaffolding Coordinates to Promote Vision-Language Coordination in Large Multi-Modal Models},
  author={Lei, Xuanyu and Yang, Zonghan and Chen, Xinrui and Li, Peng and Liu, Yang},
  booktitle={arXiv preprint arXiv:2402.12058},
  year={2024}
}

@inproceedings{zhang2024text,
  title={Text-to-Vector Generation with Neural Path Representation},
  author={Zhang, Peiying and Zhao, Nanxuan and Liao, Jing},
  journal={ACM Transactions on Graphics},
  volume={43},
  number={4},
  pages={1--13},
  year={2024},
}

@inproceedings{entem2018structuring,
  title={Structuring and Layering Contour Drawings of Organic Shapes},
  author={Entem, Even and Parakkat, Amal Dev and Cani, Marie-Paule and Barthe, Lo{\"\i}c},
  booktitle={Proceedings of the Joint Symposium on Computational Aesthetics and Sketch-Based Interfaces and Modeling and Non-Photorealistic Animation and Rendering},
  pages={1--14},
  year={2018}
}
